%% file: main.tex
\documentclass[sigconf]{acmart}

\usepackage{booktabs} 

\RequirePackage{amsfonts,latexsym,graphics,amssymb}
\RequirePackage{amsmath,amstext,url}
\usepackage{fullpage}
\usepackage{ifpdf}
\usepackage[skip=4pt]{caption}

\usepackage{amsmath,epsfig,caption,subcaption,multicol}
\usepackage{amssymb}
\usepackage{xspace}
\usepackage{ifthen}
\usepackage{array}
\usepackage{enumerate}
\usepackage{amscd}
\usepackage{xcolor}
\usepackage{color,graphicx}
\usepackage{float}

\usepackage{pbox}
\usepackage[noend]{algorithmic}
\usepackage{algorithm}

\newcommand{\M}{\mathcal{M}}
\newcommand{\I}{{\cal{I}}}

\newcommand{\Fbb}{{\mathbb{F}}}

\newcommand{\Lcal}{\mathcal{L}}
\newcommand{\Lbb}{\mathbb{L}}

\newcommand{\Ical}{{{\I}}}

\renewcommand{\I}{{\rm{I}}}

\setcopyright{none}

\begin{document}
\title{Interpreting Complex Regression Models}\thanks{${^\dagger}$The work was done while the  author was working at Yahoo Research.}

\author{Noa Avigdor-Elgrabli$^*$, Alex Libov${^\dagger}$, Michael Viderman$^*$, Ran Wolff$^*$}
\affiliation{
$^*$Yahoo Research, Haifa, Israel, 
\{noaa,viderman,ranw\}@oath.com \\ 
$^{\dagger}$Amazon Research, Haifa, Israel, alibov@amazon.com. }
\orcid{1234-5678-9012}
\renewcommand{\shortauthors}{Avigdor-Elgrabli et al.}


\begin{abstract}
Interpretation of a machine learning induced models is critical for feature engineering, debugging, and, arguably, compliance. Yet, best of breed machine learning models tend to be very complex. This paper presents a method for model interpretation which has the main benefit that the simple interpretations it provides are always grounded in actual sets of learning examples. The method is validated on the task of interpreting a complex regression model in the context of both an academic problem -- predicting the year in which a song was recorded and an industrial one -- predicting mail user churn. 
\end{abstract}

%
%




\maketitle

\input{intro.tex}
\input{problemDef.tex}

\input{algorithm.tex}
\input{useCases.tex}

\input{conclusions.tex}

\clearpage

\bibliographystyle{ACM-Reference-Format}
\bibliography{bibList} 

\end{document}

%% file: intro.tex
\section{INTRODUCTION}
Machine learning models are often divided to \emph{interpretable} and \emph{non-interpretable} (see \cite{Rudin:2014:AIM:2623330.2630823} for example). It is well accepted that simple models, such as $k$-nearest neighbors, linear classifiers and regressors, decision rules, and decision-trees can be interpreted by a professional. It is also well agreed that neural-networks, non-linear support-vector machines, and decision forests are hard to interpret. Informally, interpretability  means that a professional is able to explain the prediction on given examples, as well as the relative importance of different features.     

Proponents of using interpretable models often point to their importance as part of the broader data mining process: When feature engineering is a prolonged and iterative process \cite{anderson2013brainwash} the efficient prioritization of future efforts depends on understanding the predictions of the current model on key examples. Additionally, acknowledging that the data mining process is fallible requires debugging and trust building measures \cite{RibeiroSG16,angelino2017learning}. The establishment of trust is often much easier if stakeholders can understand why the model predicts certain outcomes. 
Last, it was argued \cite{wallace17} that modern privacy regulation such as the General Data Protection Regulation has introduced a "right to explanation" of algorithmic decisions, which is hard to satisfy without model interpretability.

Opponents of using interpretable models answer those arguments with one decisive argument: Whichever model works best in the given machine learning task is the one which should be used. The frequent superiority of complex, hard to interpret, models in common machine learning tasks is the only reason those are used. All other considerations are secondary to performance at best or mare myth \cite{ZCL2016Mythos} at worst. The recent break-through performance of deep-learning techniques, as well as their fast development cycles, seem to have concluded the argument.  

Regardless that complex  models are becoming standard, the need to prioritize feature engineering and to gain trust in model prediction remains. Thus, the problem becomes one of interpreting a complex model. That is,  gaining a measure of understanding about the main reasons a certain prediction has been given and an understanding of simple relations between features and labels. 

One way by which a prediction can be interpreted is using \emph{sensitivity analysis} \cite{saltelli2000sensitivity}. I.e.,  by measuring the effect of small variations in feature values on the prediction. The main problem with sensitivity analysis is that it requires an understanding of the features. Specifically, the meaning and legitimacy of "small" changes differ between the features. For instance, what might be a "small" change in a gender feature? One way around this problem, practiced by Robnik-\u{S}ikonja and Kononenko \cite{robnik2008} is to investigate the change in the output when the feature is missing altogether. In a recent paper, Tolomei et al. \cite{tolomei2017interpretable}  address this problem by making use of naturally occurring variations in feature values among the learning examples when the underlying model is tree-based. Both these methods interpret a model in terms of individual examples and therefore ignore the distribution of variation in a population. Therefore, they run the risk of relating exceeding importance to rare variations in feature values just because those variations influence the model prediction.

Ribeiro at el. \cite{RibeiroSG16} go beyond explaining a model using single features by develping an algorithm, LIME, which finds regions of the feature space in which the prediction of the model can be approximated by a linear model. Since a linear model is considered interpretable, the combination of a region and a suitable linear model is a good explanation. LIME still suffers the disadvantage of ignoring the probability, or even the possibility, of variation in the feature distribution. Biran and McKeown \cite{biran2017} suggest considering both the effect of a feature and its importance, which is its average effect on the set of training examples. However, their work is limited to linear models. 

We propose a model-agnostic method which interpret a model in terms of relations between events -- distinct sets of examples -- rather than single examples. Our method takes into account the actual probability of an event in the set of training examples. Hence, the explanations are not only aspects of the model but also of the population. 
For example, consider a model that predicts a user convergence probability. Our interpretation enables us to find explanations such as "the set of users that predicted to have high convergence probability has a lower proportion of females than the complementary set". The interpretations are therefore always grounded in real sub-populations of examples.


We implement our approach on two real regression problems: One is the explanation of a regression model which predicts the publication year of a song from the public million song dataset \cite{Bertin-Mahieux2011} (Section~\ref{sec:usecase1}). The other is the explanation of a user churn prediction model which is built from proprietary data (Section~\ref{sec:usecase2}). In both instances, we address the machine learning problem with a gradient boosted decision forest \cite{GBDT,XGBoost} with dozens of trees. We demonstrate important inferences about the relation of certain features to the predicted label even though the model we interpret is complex. 

%% file: problemDef.tex
\section{PROBLEM DEFINITION}
Consider a supervised learning problem. Let $F$ be a set of possible features with values from domain $\Fbb$ and let $\mathbb{L}$ be the domain of the label.
A learning example is a pair of a feature values vector $\bar{f}\in \Fbb^{|F|}$ and a label $\ell\in \mathbb{L}$ denoted by $\left\langle \bar{f},\ell\right\rangle$. Let ${\mathcal{M}}:\Fbb^{|F|}\rightarrow\mathfrak{L}$
be a model which maps every example to a distribution $\mathcal{L}$ 
on $\mathbb{L}$. We assume the model is trained on a set $A$ of
learning examples by an algorithm whose objective is to maximize the probability of correct prediction $\frac{1}{\left|B\right|}\sum_{\left\langle \bar{f},\ell\right\rangle \in B}Pr_{\mathcal{M}\left(\bar{f}\right)}\left(\ell\right)$,
where $B$ is a test set of yet unseen samples. 

We wish to describe the model $\mathcal{M}$ in simple terms. As any function, $\mathcal{M}$ can be described in
terms of its partial derivatives in selected points. However, such derivatives often are meaningless if interpreted as
variations in the features of a single example. Hence, we define the derivative of the distribution of labels in the distribution
of examples. E.g., while it is unnatural to speak of the derivative of the label with respect to the user's gender it is natural to
speak of the proportion of males in a population.

Let $T$ be an i.i.d sample of examples.  
Let $L$ be a class of distributions in $\mathfrak{L}$ and let $\overline{L}$ be its complement.  Given a model $\mathcal{M}$, let $T_{L}^{\mathcal{M}}$ be the subset of examples $\mathcal{M}$ projects into $L$, i.e., $T_{L}^{\mathcal{M}}=\left\{ t\in T:\mathcal{M}\left(t\right)\in L\right\}$ and $T_{{\overline{L}}}^{\mathcal{M}}=\left\{ t\in T:\mathcal{M}\left(t\right)\in{\overline{L}}\right\} $, respectively. 

Given a set of examples $T$, we let $\mathcal{F}_{T}$ be the empirical distribution of their features. 
Hence $\mathcal{F}_{T_{L}^{\mathcal{M}}}$ is the empirical distribution 
of the features of $T_{L}^{\mathcal{M}}$ 
and $\mathcal{F}_{T_{\overline{L}}^{\mathcal{M}}}$ is the respective empirical distribution defined by $\overline{L}$. We seek to describe 
$\mathcal{M}$ in terms of the differences between $\mathcal{F}_{T_{L}^{\mathcal{M}}}$
and $\mathcal{F}_{T_{\overline{L}}^{\mathcal{M}}}$. 

Measurement of the difference between empirical distributions is an extremely well studied problem. Some examples for such measurements are the Jensen-Shannon divergence, the Bhattacharyya distance, the Student's \emph{t} statistics and its related \emph{t}-test, and more. In our implementation we 
focus on measurement of difference between marginal distributions of single features mainly in order to further simplify  the description and to reduce measurement noise in what otherwise can be sparse data. We chose to use the 
Student's \emph{t} mainly because it can be easily computed on large data and because it is directional. 

Recall that ${\mathcal{F}}_{T}$ is the empirical distribution of the features of $T$. For a feature subset indexed by $J$ the marginal distribution of these features of $T$ is denoted by ${\mathcal{F}}_{T}^{J}$. In particular, for $j\in J$ we denote by ${\mathcal{F}}_{T}^{j}$ the marginal distribution on the $j^{th}$ feature. 

In this paper, instead of $T$ we often consider a subset of examples $T_{L}^{\mathcal{M}}$ and its complement $T_{\overline{L}}^{\mathcal{M}}$. With some abuse of notations, a projection of these examples on a feature subset (marginal) $I \subseteq F$, defines marginal distribution   
$\Ical_{T_{L}^{\mathcal{M}}}$ and $\Ical_{T_{\overline{L}}^{\mathcal{M}}}$, respectively.  

Now, we are going to formally define the model interpretation problem. Then, Example~\ref{ex:dismeasure} will provide an example of the concepts defined in this section. 

\begin{definition}\label{def:modelinter}
Given a model $\mathcal{M}$, a set of examples $T$, and a measure $dis$ 
for distribution dissimilarity, the model interpretation problem
is to select a class of distributions $L$ out of a family of such classes, and a marginal $I$ out of a family of marginals such that 
$dis\left(\Ical_{T_{L}^{\mathcal{M}}},\Ical_{T_{\overline{L}}^{\mathcal{M}}}\right)$
is maximized. 
\end{definition}

Now let us present a simple example demonstrating the concepts we defined earlier. 

\begin{example}[Distribution dissimilarity measure]\label{ex:dismeasure}
Assume a set of features $F = \left\{f_1, f_2 \right\}$ and 
a set of examples 
$$T = \left\{t_1=(1,2),t_2=(3,4),t_3=(1,3), t_4=(5,6)\right\},$$
where every example $t_i$ is associated with a tuple of $f_1$ and $f_2$ feature values. 
Let the labels domain be $\mathbb{L} = \{1,2,3,4\}$, and assume that 
$\mathcal{M}\left(t_1\right) = 1$, $\mathcal{M}\left(t_2\right) = 2$, $\mathcal{M}\left(t_3\right) = 3$, $\mathcal{M}\left(t_4\right) = 4$ with probability 1. 

Then, considering feature $f_1$ we have a marginal distribution ${\mathcal{F}}_{T}^{1}$ defined by the sample $\left\{1,3,1,5\right\}$. With some abuse of notations let 
$L = \{1,2\}$ and then $\overline{L}=\{3,4\}$. Notice that the provided subsets $L$ and $\overline{L}$ is only a one out of multiple options to split the labels domain $\mathbb{L}$. 
In this case, $T_{L}^{\mathcal{M}} = \{t_1, t_2\}$ and $T_{\overline{L}}^{\mathcal{M}} = \{t_3, t_4\}$. 

Then, considering a projection of examples on feature $f_1$, we can apply a measure $dis$ as follows 
$$dis\left(\left(f_1\right)_{T_{L}^{\mathcal{M}}}, \left(f_1\right)_{T_{\overline{L}}^{\mathcal{M}}}\right) = dis\left(\left(f_1\right)_{\{t_1, t_2\}}, \left(f_1\right)_{\{t_3, t_4\}}\right) = $$
$$dis\left(\{1,3\}, \{1,5\}\right).$$ 

To solve the model interpretation problem (Definition \ref{def:modelinter}), i.e., to maximize the $dis$ measure, one needs to consider the projections on $f_1$, $f_2$ and all   possible options to split the labels domain $\mathbb{L}$ to $L$ and $\overline{L}$. 
\end{example}

Specifically, this paper concerns with a problem in which the label domain is $\mathbb{R}$ and the family of distributions is $L_{\left[a,b\right]}= \left\{N\left[\mu,\sigma^2\right] : \mu \in \left[a,b\right]\right\}$. The domain of all features is numeric except for missing values, $\mathbb{F} = \mathbb{R}\cup \left\{\emptyset\right\}$. 
Interpretations are single feature $F^{j}$  with marginal distribution $\mathcal{F}_{T_{L}^{\mathcal{M}}}^{j}$ and the dissimilarity measure is the Student's \emph{t}-test. 

To ease notation, for the rest of the paper we will refer to each label distribution $L_{\ell}=\left\{N\left[\mu,\sigma^2\right] : \mu = \ell \right\}$ by its mean value $\ell\in L$. We will refer to each class of label distributions $L_{[a,b]}$ by its mean value range $[a,b]$ (when referring to a set of examples). We denote by $[a_{min},a_{max}]$ the overall class of distributions range received by applying $\M$ on all examples in $T$. As $\M$ and $T$ are fixed we will denote $dis\left(\hat{\I}_{T_{L}^{\mathcal{M}}}^{j},\hat{\I}_{T_{\overline{ L}}^{\mathcal{M}}}^{j}\right)$ by $dis(I^j,L)$ (we also use $dis$ when $I^j$ and $L$ are clear from the context).

%% file: algorithm.tex
\section{Algorithm}
The number of potential interpretations of a model is linear in the number of features and quadratic in the number of unique predicted label values. In a complex regression model, it is common that the number of unique predicted label values would be proportional to the number of examples. Furthermore, since many of the features can be sparse, many examples are often needed to accurately measure the dissimilarity of values in-segment vs. out-of-segment feature values. Searching for the most informative set of descriptions is therefore tasking in terms of runtime performance.  

\begin{figure}[htbp!]
\includegraphics[width=0.45\textwidth]{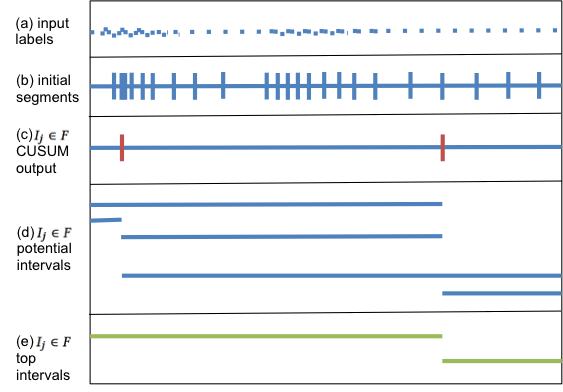}
\caption{Graphic representation of the algorithm process. (a) Illustration of the input labels distribution. (b) The examples partition into small segments each with enough examples.
(c) The CUSUM algorithm output given the sequence of the $dis$ evaluation of each segment. (d) Illustration of all possible bigger segments. (e) The top two segments in terms of the dissimilarity measurement.
Phase (b) is done regardless of the features values, phases (c)-(e) are executed for each feature separately.}\label{fig:alg_schem}
\end{figure}

\begin{figure*}[h!] 
\centering
\captionsetup{justification=centering}
\begin{subfigure}[b]{0.48\textwidth}
    \centering
	\includegraphics[width=\textwidth]{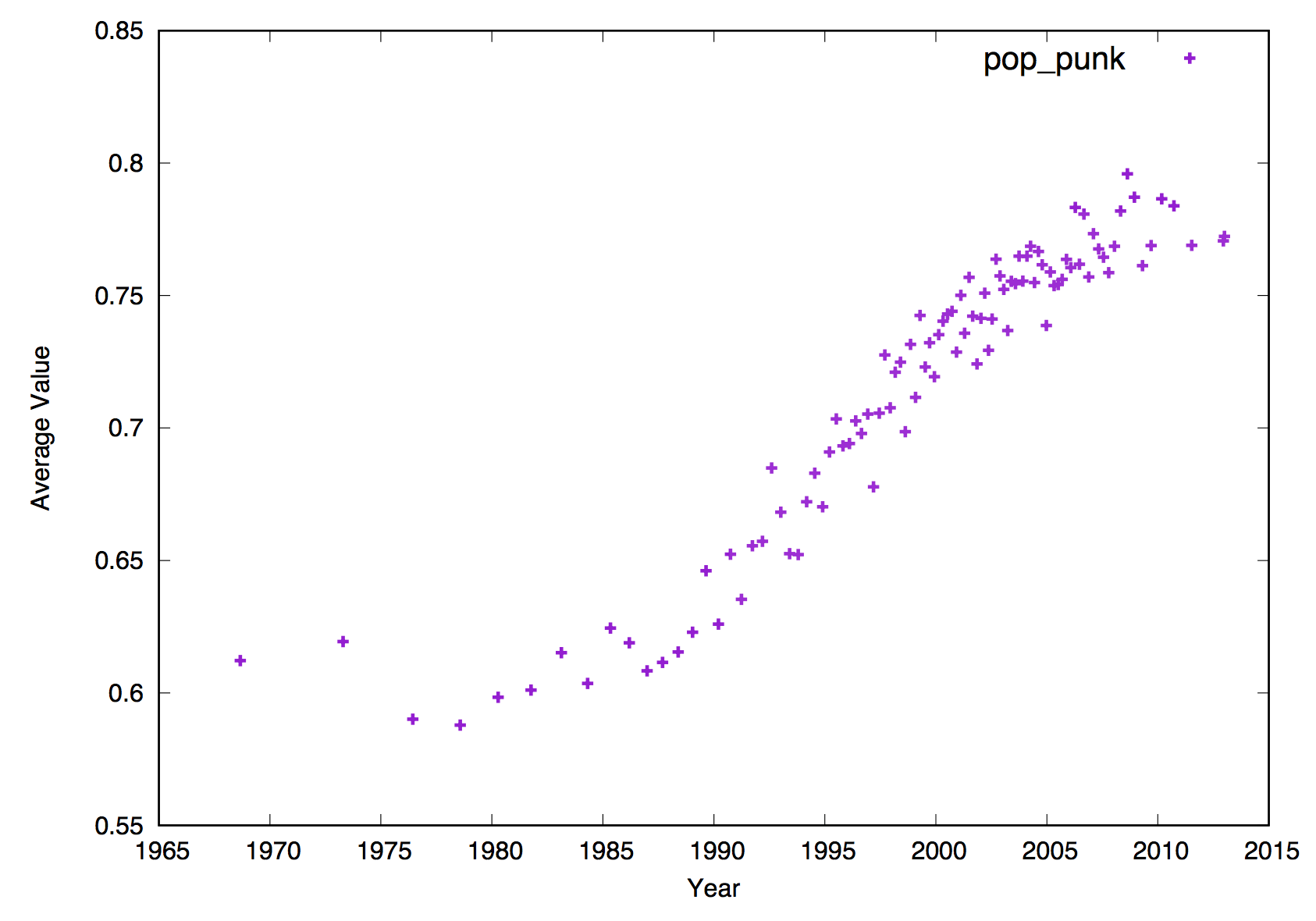}
        \vspace{-0.1in}
        \hspace{-0.1in}
    \caption{average value per initial segment\label{fig:MSD_pop_punka}}%
\end{subfigure}
\begin{subfigure}[b]{0.48\textwidth}
	\centering
	\includegraphics[width=\textwidth]{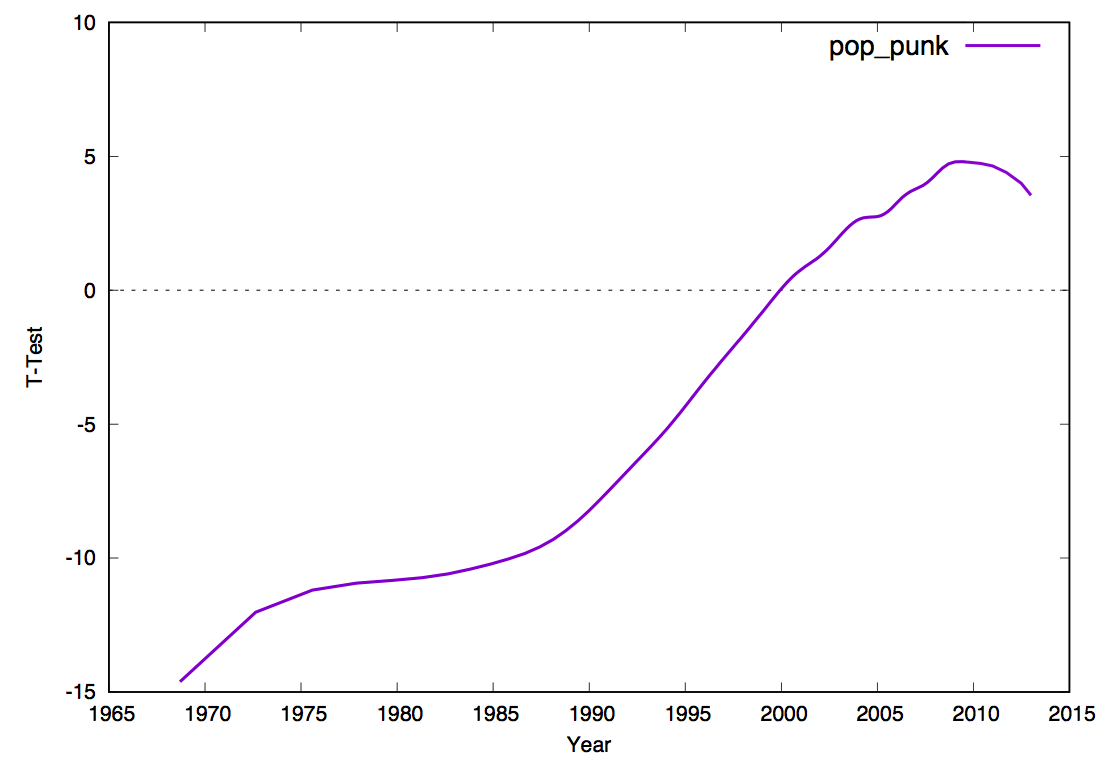}
        \vspace{-0.1in}
        \hspace{-0.1in}
         \caption{\emph{t}-test per initial segment\label{fig:MSD_pop_punkb}}%
\end{subfigure}

\begin{subfigure}[b]{0.48\textwidth}
	\centering
	\includegraphics[width=\textwidth]{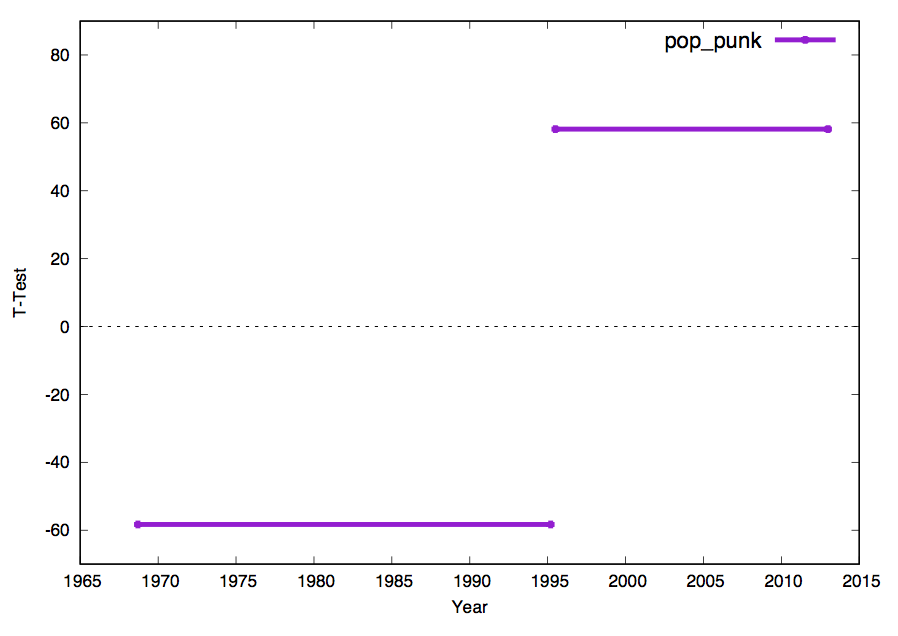}
        \vspace{-0.1in}
        \hspace{-0.1in}
         \caption{selected segments\label{fig:MSD_pop_punkc}}%
\end{subfigure}
\caption{`Pop punk' feature: an example to the segmentation process}
\label{fig:MSD_pop_punk}
\vspace{-0.1in}
\end{figure*}

This section describes a three steps algorithm which efficiently selects a set of interpretations for a model. The first step of the algorithm is to measure feature dissimilarity in and out of small bins of the label space. The second step considers the dissimilarity measurement as a random variable and searches, using an efficient linear algorithm,  for larger segments in which the random variable is stationary. This greatly reduces the complexity of the problem. Then, at a third step, those larger segments scored, ranked, and clustered to produce the final outcome. The pseudo-code of the algorithm is described in Algorithm~\ref{alg:finalalg} and a graphic representation of the process can be found in Figure~\ref{fig:alg_schem}. A further illustration of the segment selection process can been seen in Figure~\ref{fig:MSD_pop_punk}, showing the analysis phases done on one feature (see Section~\ref{sec:usecase1} for more details).

\begin{algorithm}
\caption{Model interpretation algorithm}\label{alg:finalalg}
 Given example set $T$ over a feature set $\Fbb$ with labels in the range $\Lbb=[a_{min},a_{max}]$, and given parameters $t,k,m \in \mathbb{N}$.
\begin{enumerate}

 \item {\bf Initial binning}
 \begin{enumerate}
 \item Partition $[a_{min},a_{max}]$ into $k$ subranges $[b_0,b_1)$, $[b_1,b_2)$,$\ldots$, $[b_{k-1},b_k]$ as follows: 
  \begin{enumerate}
 \item Pick sample $S \subseteq T$ s.t. $|S| = 2m\cdot k$, and $\forall l\in\Lbb $ $|\{e\in S:l_e=l \}| \le m  $. \noindent
 \item $b_0\leftarrow a_{min}$, $b_k\leftarrow a_{max}$, else $b_i\leftarrow l_{S[2m \cdot i -1]}$ ($S$ is in descending order)
\end{enumerate}
 \item $\forall I_j\in \Fbb$, $\forall i\in [k]$:\newline
 set ${\mathcal{B}}_j(i)=dis\big(I_j,[b_i,b_{i+1})\big)$
\end{enumerate}
 \item {\bf Identify potential segments}
 \begin{enumerate}
\item  $\forall I_j\in \Fbb$ apply standard normalization\indent\newline
${\mathcal{B}}'_j(i)\leftarrow \frac{{\mathcal{B}}_j(i)-\mu}{\sigma}$ ($\mu=Mean({\mathcal{B}}_j)$, $\sigma=STD({\mathcal{B}}_j)$ )
\item $C \leftarrow$ CUSUM(${\mathcal{B}}'_j$) $\cup \{a_{min},a_{max}\}$
\item $P \leftarrow \{[i_1,i_2] : i_1<i_2 \in C\}$ 
\item Order $P$ in descending order of the $dis$ value
\item Set $Int(I_j)$ as follows: Add $Int(I_j)$ segments from $P$ that doesn't intersect previously picked segments.
 \end{enumerate}
 \item {\bf Output $t$ top segments}
 \begin{enumerate}
\item $Int\leftarrow \bigcup_{I_j\in\Fbb} Int(I_j)$ 
\item Order $Int$ in descending order of the $dis$ value
\item Set $Top$ to be first $t$ segments from $Int$. \\
      If required, $Top$ can be selected only over a predefined subset of features. 
\item Output $Top$ 
 \end{enumerate} 
  
 \item {\bf Cluster segments}
 \begin{enumerate}
\item $Int\leftarrow \bigcup_{I_j\in\Fbb} Int(I_j)$ 
\item Cluster $Int(I_j)$ to $C_1,\ldots,C_k$
\item Set $Repr_k$ to be the set of $k$ segments, where a segments with highest $dis$ value is taken from each $C_j$ for $j\in \{1,\ldots,k\}$. 
\item Order $Repr_k$ in descending order of the $dis$ value
\item Set $Top_{cluster}$ to be first $t$ segments from $Repr_k$.
\item Output $Top_{cluster}$ 
 \end{enumerate} 

\end{enumerate}
\end{algorithm}

 \subsection{Step 1: Initial binning}\label{sec:featrepr}

The first step of the algorithm is to divide the label space into small segments (bins) which are just large enough to compute the dissimilarity of feature values. We denote the $i^{th}$ bin by $\left[b_{i},b_{i+1}\right)$ and set the values of $b_{i}$ by evenly dividing a large sample.  Taking into account that some features may be sparse we aim to have for each feature and every bin at least one sample whose value is not missing.  The number of segments is denoted by $k$ and the number of samples in an average segment by $m$. See Figure~\ref{fig:alg_schem}(b) for a graphic representation.

For each feature $I_j\in F$ and every segment we calculate their distributional dissimilarity. We henceforth denote ${\mathcal{B}}_j(i)=dis(I_j,[b_i,b_{i+1}))$ the value of the dissimilarity function for feature  $I_j\in F$ on segment $\left[b_{i},b_{i+1}\right)$. The dissimilarity measure we used in our implementation is Student's two samples $t$.\footnote{We experimented with other measures, such as the Kullback--Leibler divergence but saw no systematic difference.} 

\begin{definition}
Given two samples from distributions $\Lcal_1$, and $\Lcal_2$. For $i\in \{1,2\}$, let $\mu_i,  \sigma_i, n_i$ denote the $\Lcal_i$ sample average, standard deviation, and number of elements, respectively.
The (student) $2$-sample t-test statistic is 
\[ \frac{\mu_1-\mu_2}{\sqrt{\frac{\sigma_1^2}{n_1} + \frac{\sigma_2^2}{n_2} }} \]
\end{definition}

\subsection{Step 2: Identify Potential Segments}\label{sec:findinter}
Given the initial binning, the second step of Algorithm~\ref{alg:finalalg} is to compute potential segments which may maximize the dissimilarity. For that, we use the CUSUM  change point detection algorithm \cite{page54, Hink71}. 
CUSUM detect, with some probability of an error, segments of bins in which the differences between the values of ${\mathcal{B}}_j(i)=dis(I_j,[b_i,b_{i+1}))$ are due to random variation, rather than an actual change. The benefit of using CUSUM are that it is linear in the number of observations (bin) and yet asymptotically optimal in terms of the number of errors.

CUSUM has one parameter which is the size of changes which should be ignored. To be able to use the same parameter value with features whose ${\mathcal{B}}_j(i)$ can vary we first normalize the values of every feature using z-score (i.e., subtract their mean and then divide by their variance). The normalized sequence is denoted by ${\mathcal{B}}'_j(i)$. The output of CUSUM is a set of bins $b_j(i_{1}),b_j(i_{2}),\ldots b_j(i_{\ell})$ in which the value of  ${\mathcal{B}}'_j(i)$ seems to have gone a systematic change. This set (see Figure~\ref{fig:alg_schem}(c))  necessarily holds many fewer than $k$ bins. 

The output of CUSUM is a set of change points. Dissimilarity measure such as Student's $t$ increases with the sample size as long as there is no change in the distribution. However, the dissimilarity may still increase even if there has been a change (specifically if the change is minor comparing to other segments). Therefore, a potential segment maximizing the $dis$ measure may start and end in any of the change points detected by CUSUM. Hence, in Step 2 of Algorithm~\ref{alg:finalalg} all potential segments are generated.  Namely if $C$ contains all change points, we set $P \leftarrow \{[i_1,i_2] : i_1<i_2 \in C\}$ to contain the start and the end locations of every potential segment (see Figure~\ref{fig:alg_schem}(d)). After that we pick the segments having highest $dis$ values.  



\subsection{Steps 3 and 4: Clustering and selection}\label{sec:bestintselect}
Here we explain last two steps of Algorithm~\ref{alg:finalalg}.

\paragraph{Step 3 - Top Segments selection}

The outcome of the maximal segment selection step is a small set of segments, for every feature, which maximizes the in-segment vs. out-of-segment dissimilarity of that feature. Yet the set of features is usually very large and features, as well as their dissimilar segments, tend to correlate. The $3^{rd}$ step of the algorithm therefore selects a set of interesting and representative segments which should be presented to the user. 

Unlike the previous steps, whose outcome is defined in terms of optimizing a statistics, this stage of Algorithm~\ref{alg:finalalg} optimizes the user experience. The simplest criteria by which segments are selected for presentation is their dissimilarity. We take advantage of the fact that the Student's $t$ statistics scores all features on a joint scale, and use it as a scoring function. Hence, one view of the output is the top ten (or any other constant) most dissimilar segments. 

Often, not all features are equally important. For instance, we find that some users are only interested in a subset of the features which they understand. Other users are only interested in aspect which they can control. A second option is therefore to rank only the segments of a subset of these features.

\paragraph{Step 4 - Clustering}
Last, features are often strongly dependent. Thus, we find it useful to cluster the segments according to their important characteristics: Beginning, end, sign of the Student's $t$ statistics, and the textual similarity of the feature name. We use $k$-means clustering \cite{lloyd82,forgy65} with k-means++ initial seeding \cite{kmeanplusplus2007}. Inspired by \cite{BischofLS99} we use a minimum description length (MDL) method to select the optimal value of $k$. The MDL cost of every clustering solution is taken to be the sum of the log of the size of the vector of the centroid plus the sum of the log of the distance of every segment from its nearest centroid, and we choose the $k$ which minimizes this MDL cost.

%% file: useCases.tex
\section{Experiments}

We applied the model interpretation algorithm to two regression models. The first is a model which was developed for an actual industrial use case of predicting changes in the activity of users of one of the largest e-mail service providers. The other is a model which was developed for the million songs database and which predicts a song's release year. The second data set is freely accessible.

In both instances, a gradient-boosted decision forest (GBDT)\\~\cite{GBDT,XGBoost} was trained to predict the numeric label. GBDT typically induces hundreds or thousands of trees each of whom has a numeric prediction in every leaf. The prediction of the forest is the weighted sum of the predictions of the trees. It is fair to say that GBDT is very hard to interpret directly. The most that existing packages provide is a feature importance metric which is the count of the tree splits which use every feature. However, in this statistic a feature which is used by 10\% of the splits on the $6^{th}$ level of one hundred trees would have a higher count (10\% of $2^{6} \times 100=320$) than a feature which is used in all of those trees roots.

\subsection{Interpreting a user activity change model}\label{sec:usecase2}

A large mail provider is tracking the activity of its users using multiple metrics. The provider then trains a predictive model which predicts those changes. Beyond the value of expecting the change before it occurs, the main purpose the provider has is distilling the systematic aspects of user's change in activity. However, since the predictive model is opaque, the provider requires an interpretation of the model.

\begin{figure}[htbp!]
\centering
\includegraphics[width=0.45\textwidth]{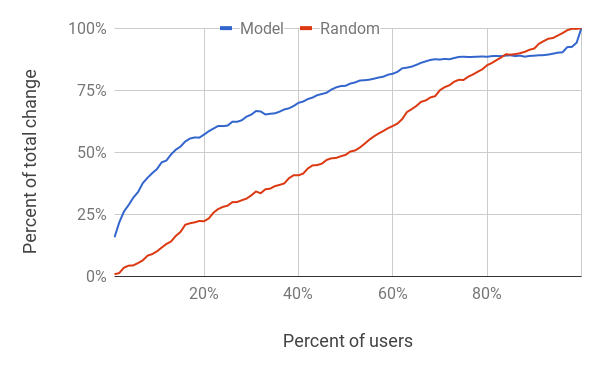}
\caption{Average change of activity as a function of the percent of the user population when ranked according to predicted risk vs. when ranked at random. The sharp increase of the average change at lower percentages indicate that the model captures those users who do change their activity.}\label{fig:changeactivity}
\end{figure}

The model we interpret is trained using data which was collected on one million users over a period of eight weeks. The features include user features (gender, age, zip code, country, state, Browser-Cookie age, device type and roaming), and user activity (user seen, counters of actions like read, send, search ,  and orgenize actions like archive, move, mark as unread, mark as Spam and delete). On top of the user activity we calculate raw counts and basic statistics like mean value and standard divination (stdDev) and more complex evaluations like the parameters of a fitted Holt-Winters (FC) time series model or the parameters of CUSUM change detection evaluation on the user's activity vector. The label is the difference of activity in a key metric between the ninth and the tenth weeks. The predicted label is perceived by the provider as that user's risk. Figure \ref{fig:changeactivity} depicts the success of the model in predicting a change in user behavior vs. a random benchmark. It can be seen that the model is able to capture about 20\% of the users who are responsible to nearly 60\% of the risk. This is despite a low Spearman correlation of 0.11 between the predicted label and the actual label.


\input{experimental-results-churn.tex}

\subsection{Interpreting a song release year model}
\label{sec:usecase1}

To allow repeating our results on an academic dataset we built and then interpreted a model for the  publicly available Million Song Dataset (MSD)~\cite{Bertin-Mahieux2011}. 
That dataset consists of nearly one million songs and a large number of features for every song. As a toy prediction task, we chose to train a model which will predict the release year of a song.
We scanned the meta parameters of GBDT by training one thousand models each with 25K examples and testing each  model on 10K disjoint examples. We then chose the model which showed the best Spearman correlation between label and predicted label, 0.204. We stress that the predictive accuracy of the model is not central to the interpretation task, although a model which produces random prediction will have no interpretation.

\input{exp_result_msdb.tex}

%% file: experimental-results-churn.tex
\begin{figure*}[!hbt]
\centering
\captionsetup{justification=centering}
\begin{subfigure}{0.48\textwidth}
	\centering
	\includegraphics[width=\textwidth]{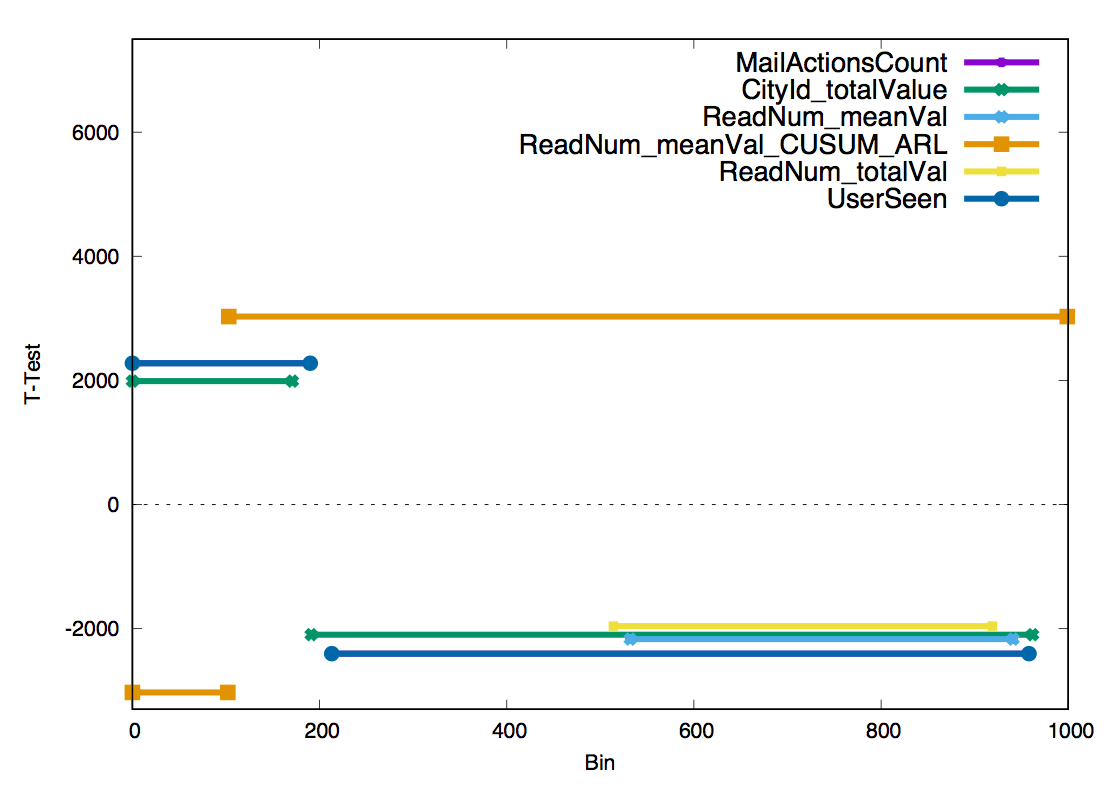}    
     \vspace{-0.3in}
    \caption{segment's \emph{t}-test per bin}\label{fig:m1}    
\end{subfigure}
\begin{subfigure}{0.45\textwidth}
\centering
\includegraphics[width=\textwidth]{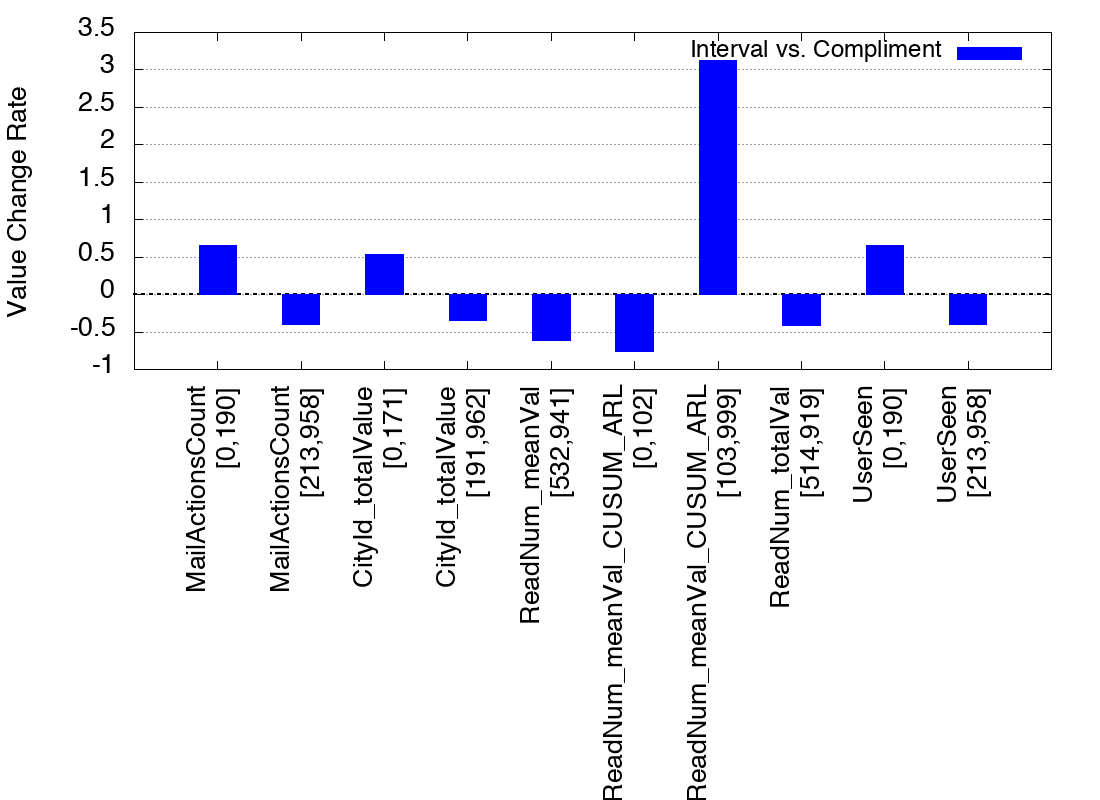}

\caption{segment's vs. complement population}\label{fig:m2}
\end{subfigure}

\caption{Top segments over all features.}\label{fig:avgcomparison}
\end{figure*}

\begin{figure*}[!hbt] 
\centering
\captionsetup{justification=centering}

\begin{subfigure}{0.48\textwidth}
	\includegraphics[width=\textwidth]{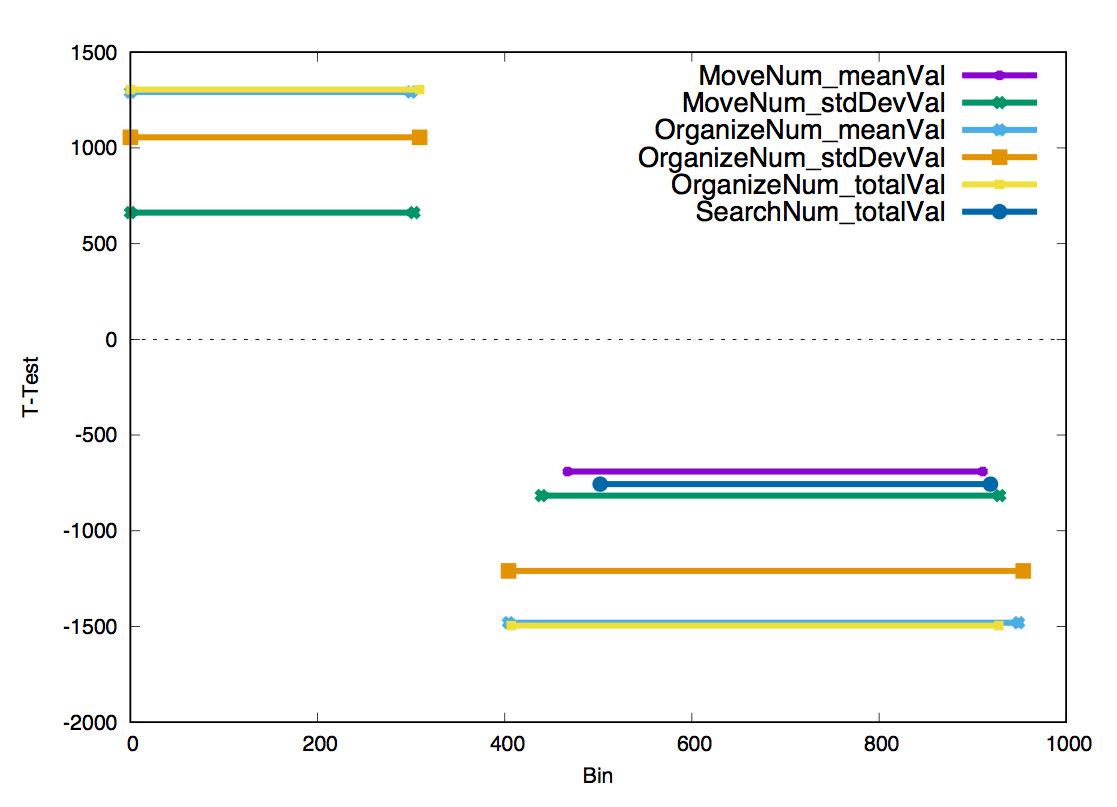}
    \vspace{-0.2in}
    \caption{advanced mail features}
\end{subfigure}
\begin{subfigure}{0.48\textwidth}
	\centering
	\includegraphics[width=\textwidth]{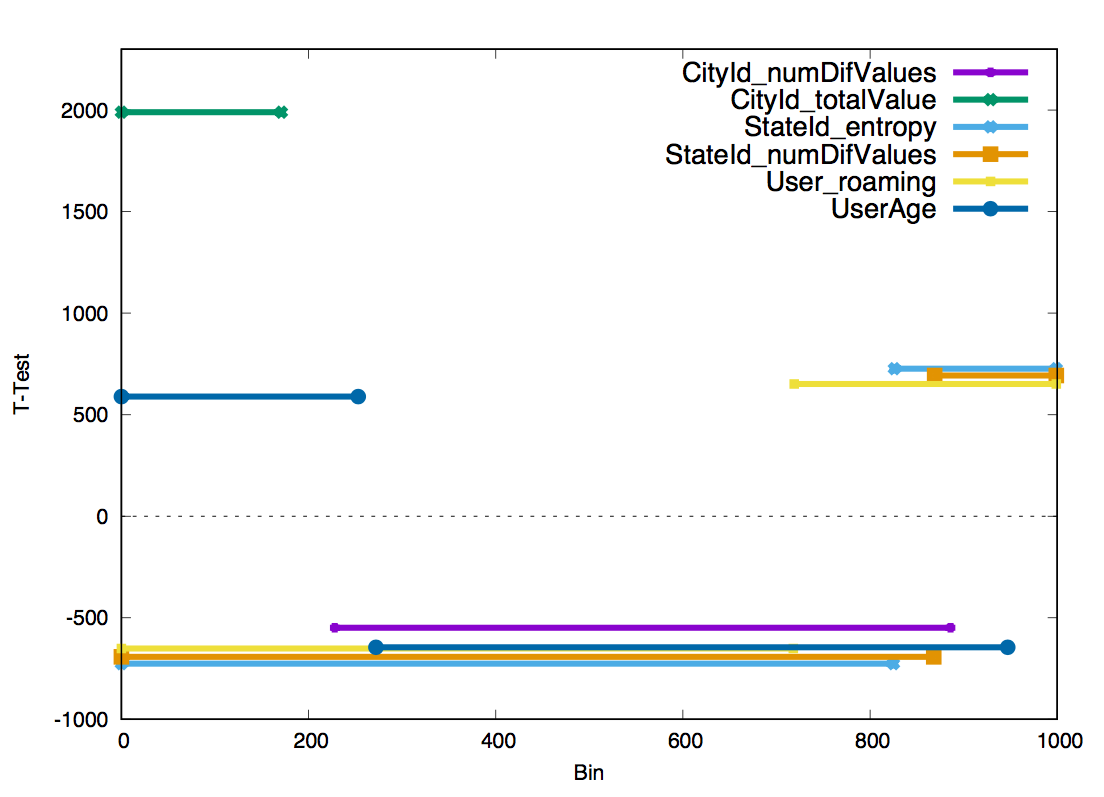}
            \vspace{-0.2in}
   	\caption{user features}\label{fig:mailuserftrs}	
\end{subfigure}

\caption{Interpretation in terms of advanced mail and user features}\label{fig:mailadvanced}
\end{figure*}

\begin{figure*}[!hbt]
\centering
\captionsetup{justification=centering}
\begin{subfigure}{0.48\textwidth}
	\centering
	\includegraphics[width=\textwidth]{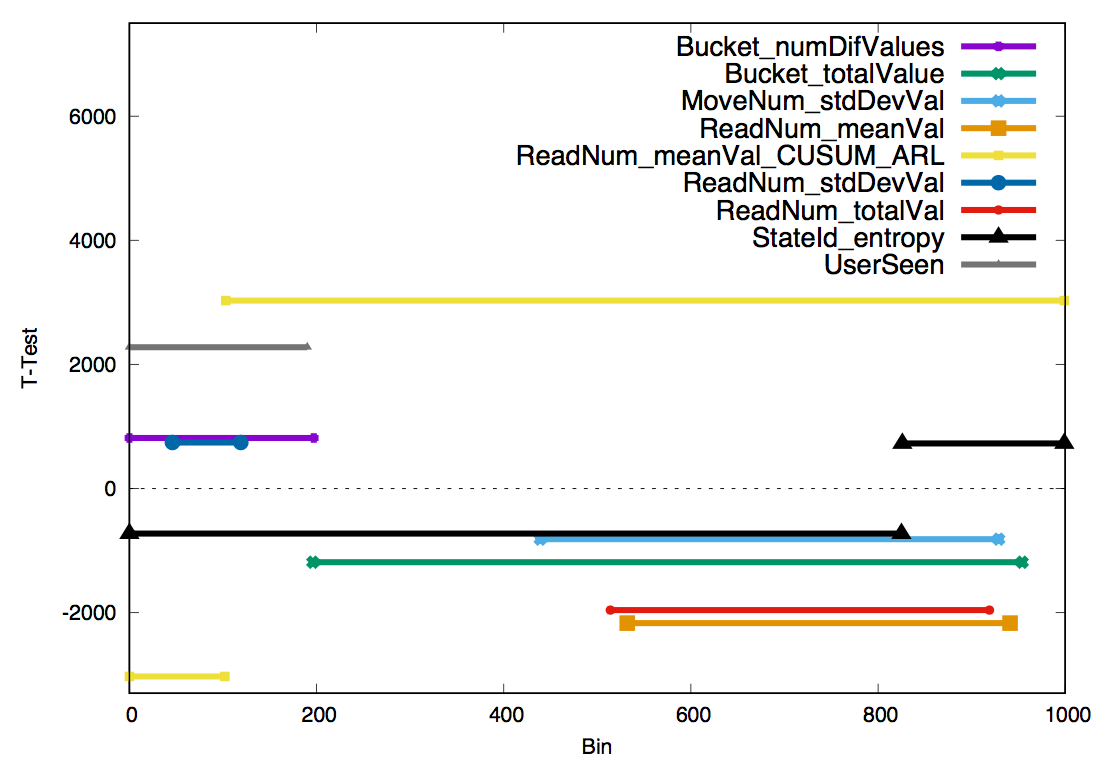}    
     \vspace{-0.2in}
    \caption{One representative from each cluster}\label{fig:clm1}    
\end{subfigure}
\begin{subfigure}{0.48\textwidth}
	\centering
	\includegraphics[width=\textwidth]{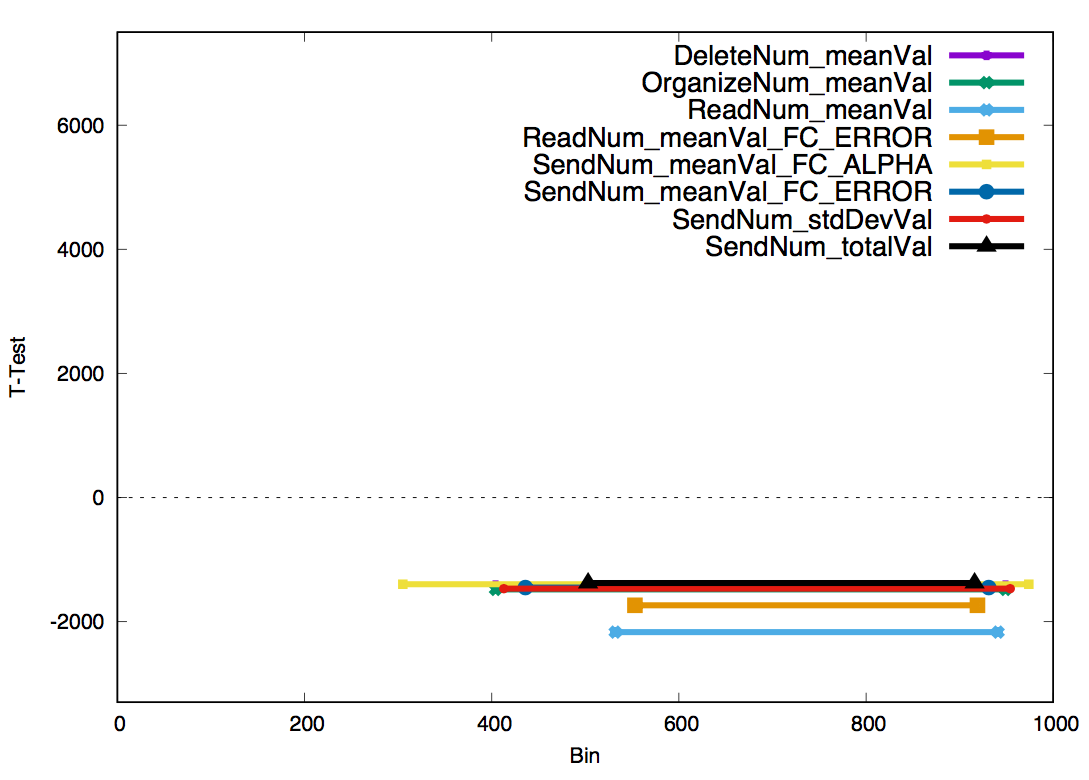}    
     \vspace{-0.2in}
    \caption{'ReadNum\_meanVal' cluster}\label{fig:clm2}    
\end{subfigure}

\begin{subfigure}{0.48\textwidth}
	\centering
	\includegraphics[width=\textwidth]{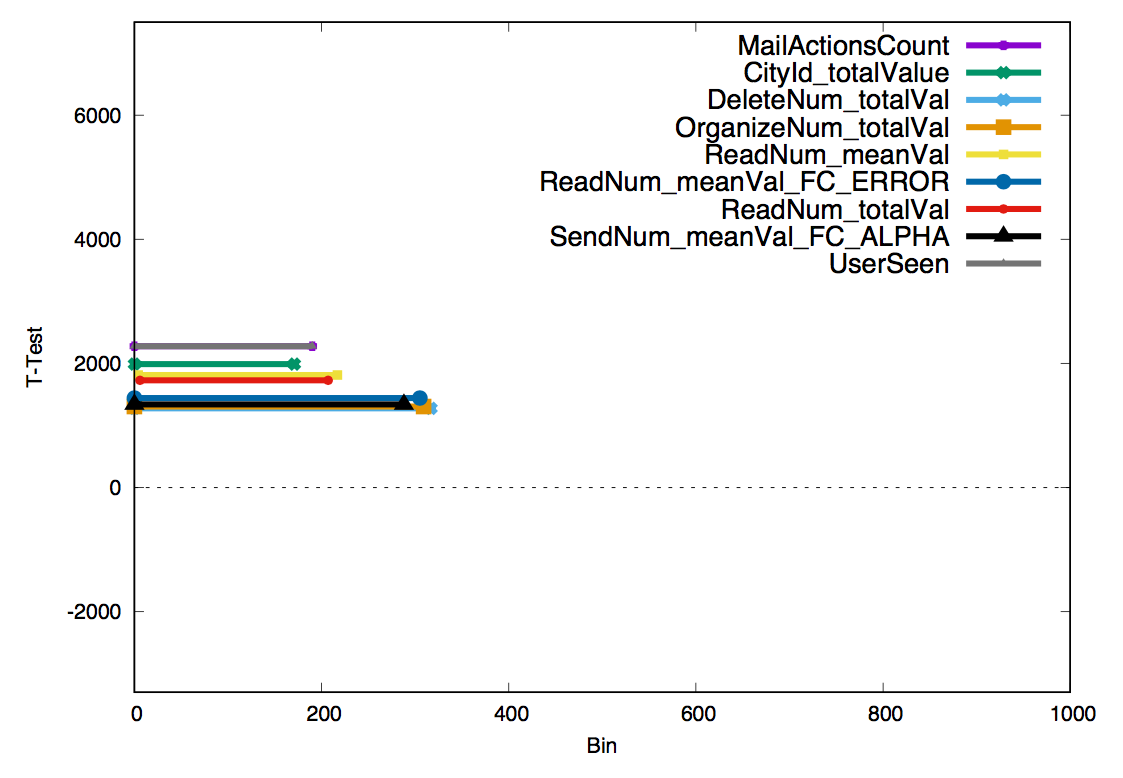}    
     \vspace{-0.2in}
    \caption{'UserSeen' cluster}\label{fig:clm3}    
\end{subfigure}
\begin{subfigure}{0.48\textwidth}
	\centering
	\includegraphics[width=\textwidth]{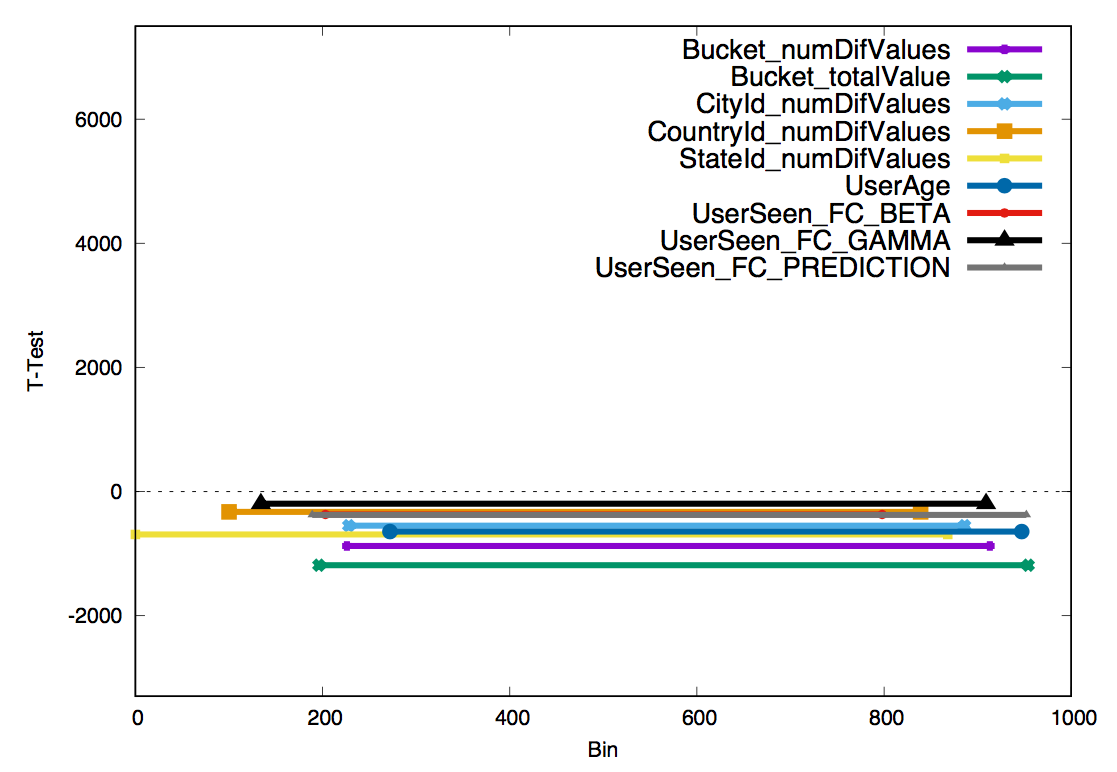}    
     \vspace{-0.2in}
    \caption{'Bucket\_totalValue' cluster}\label{fig:clm4}    
\end{subfigure}
\caption{Clustering of Segments.}\label{fig:clustinter}
 \vspace{-0.1in}
\end{figure*}

\begin{table*}[!htb]  
    \small
    \begin{subtable}{.5\linewidth}        
         \begin{tabular}[b]{ | p{4.3cm} | l | }
    \hline   
    Feature & Importance  \\ \hline
    \midrule
    ReadNum\_stdDevVal & 154 \\ \hline
    ReadNum\_meanVal & 151 \\ \hline
    ReadNum\_meanVal\_CUSUM\_ARL  & 150 \\ \hline
    ReadNum\_meanVal\_FC\_ERROR & 111 \\ \hline
    ReadNum\_meanVal\_FC\_PREDICTION & 88  \\ \hline
    UserSeen\_FC\_ALPHA & 85  \\ \hline
    UserFeature\_bcookieAge\_max & 84  \\ \hline
    UserSeen\_FC\_ERROR & 80 \\ \hline
    UserAge & 79 \\ \hline
    ReadNum\_meanVal\_FC\_ALPHA & 77 \\
    \bottomrule
  \end{tabular}
  \caption{GBDT feature importance}\label{tab:mailfeatures1}
    \end{subtable}%
    \begin{subtable}{.5\linewidth}        
  \begin{tabular}[b]{ | p{3.7cm} | l | l | l | }
    \hline
    Feature & Segment & Student's $t$ & Importance\\ \hline
    \midrule
    ReadNum\_meanVal\_CUSUM\_ARL & [0,102] & -3030.06 & 150 \\ \hline
    ReadNum\_meanVal\_CUSUM\_ARL & [103,999] & 3030.06 & 150\\ \hline
    UserSeen & [213,958] & -2404.71 & --- \\ \hline
    MailActionsCount & [213,958] & -2404.71 & 73  \\ \hline
    MailActionsCount & [0,190] & 2276.71 & 73 \\ \hline
    UserSeen & [0,190] & 2276.71 & --- \\ \hline
    ReadNum\_meanVal & [532,941] & -2170.19 & 151\\ \hline
    CityId\_totalValue & [191,962] & -2099.33 & 23 \\ \hline
    CityId\_totalValue & [0,171] & 1989.84 & 23 \\ \hline
    ReadNum\_totalVal & [514,919] & -1958.32 & 72\\
    \bottomrule
  \end{tabular}
  \caption{Strongest interpretations}\label{tab:mailfeatures2}
    \end{subtable} 
    \caption{Feature importance comparison - User Model\label{tab:mailfeatures2}}
\end{table*}

We executed the interpretation model with $k=1000$ initial bins where the lower (left) bins correspond with the sharpest decrease in activity and the higher (right) with the sharpest increase. Because of business secrecy we cannot share actual numbers in this use case and instead we report percentiles. Figure \ref{fig:avgcomparison}(a) depicts the ten most dissimilar segments the algorithm has identified in terms of their label segment and their Student's $t$.  Figure~\ref{fig:avgcomparison}(b) depicts the ratio of the relative change of average feature value in the segment vs. out-of the segment. For instance, the feature  "ReadNum\_meanVal\_CUSUM\_ARL" which measures how unpredictable a user's reading behavior is turned out to be a very strong indication for high risk: For about the 10\% of the users whose risk is the highest that feature is three times lower than average, that is the users are less predictable. For the other 90\% the feature is on average four times  higher than average. Other features, which are perhaps more surprising are a higher number of cities visited by the user, which relates to high risk and a lower mail readership which relates to low risk.

%
This ability to uncover hidden relations is further stressed when the user requests to interpret a model in terms of a specific set of features. Figure~\ref{fig:mailadvanced}(a) depicts the most dissimilar segments of features related to advanced mail features such as searching the mailbox or organizing the mailbox. Such view is important when developing new features for the product. Our analysis revealed that higher risk users make more use of some of those advanced features, but have a higher standard deviation which might mean that they try the feature but do not attach to it. On the other hand, Figure~\ref{fig:mailadvanced}(b) shows that lower risk users modify their 'StateId' and use 'roaming' more often, which probably implies that they travel more. 

Simply sorting the explanations according to their Student's $t$ is unsatisfying because many of the shown features are related to one another. For instance, the "ReadNum\_totalCount" feature counts the number of read mails and the "MailActionsCount" feature also counts other actions but is dominated by reads. Clustering greatly increases the expressiveness of the output. Figure \ref{fig:clustinter} depicts a summary of just one representative from each cluster (Fig. \ref{fig:clustinter}(a) and then a breakout of all of the segments which where clustered together for  three specific clusters. This hybrid view better describes the way different features interact with the label and with each other.


Table \ref{tab:mailfeatures2}  compares the interpretations provided by our algorithm to GBDT feature importance report. One drawback of the GBDT features importance is that it only takes into account the features used by the GBDT model. Therefore, if there are several correlated features GBDT might choose some of them and ignore the others while in our analysis all the informative features will arise. And indeed, the most striking difference between Table \ref{tab:mailfeatures2}(a) and Table \ref{tab:mailfeatures2}(b) is that "UserSeen" (the number of active days the user had in eight weeks) have one of the most significant segments but not even listed in the feature importance report of GBDT. Apparently, when making split decisions, GBDT always found that  other features where more informative at the local level and thus the global importance of the "UserSeen" feature is overlooked in the feature importance report. 
Moreover, the order between the features by the GBDT table is determined by the number of appearance in the GBDT forest and not necessarily represent the true importance of the feature. For example, "MailActionsCount" listed as number 17 in the GBDT feature importance report (not seen in the table) and well below the "ReadNum\_meanVal" feature. The strongest interpretation report unveils that in two specific segments of the label space "MailActionsCount" is in fact much more related to the label than  "ReadNum\_meanVal". 

Lastly, the most obvious difference is that unlike our method, GBDT may list its most important features but its feature report does not contain the labels range where features has an outstanding behaviour.

%% file: exp_result_msdb.tex
\begin{figure*}[h!] 
\centering
\captionsetup{justification=centering}
\begin{subfigure}[b]{0.48\textwidth}%
	\centering
	\includegraphics[width=\textwidth]{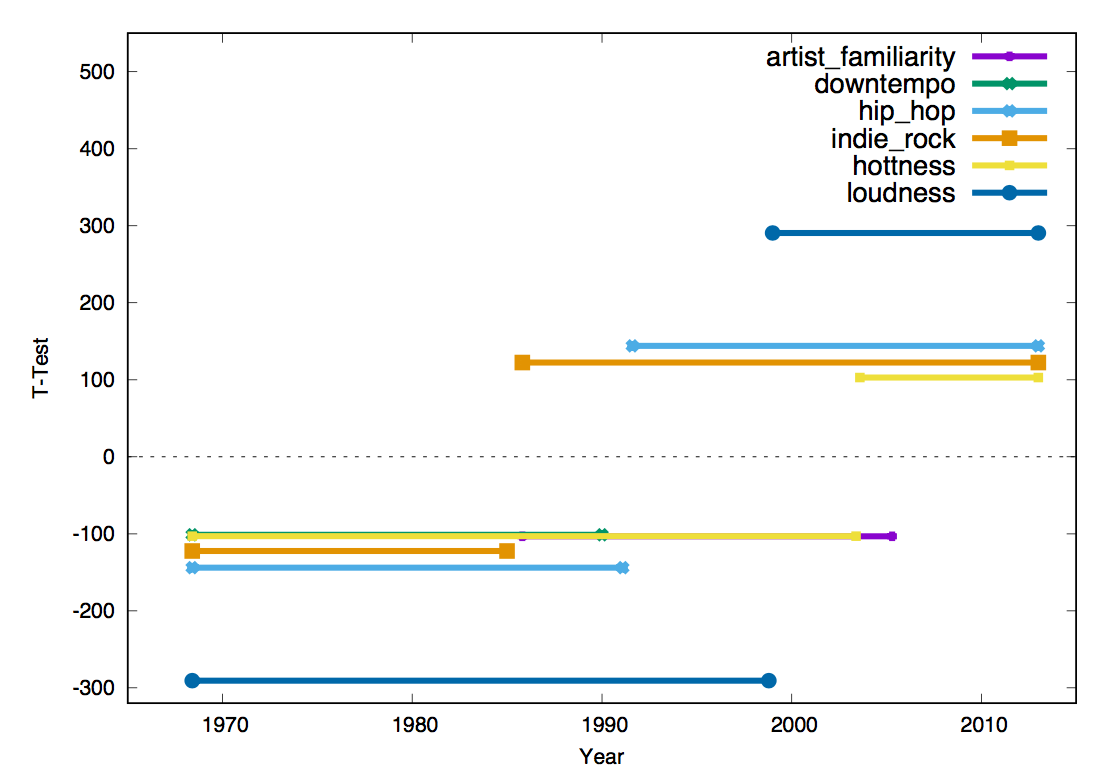}%
    \caption{Segments range and Student's $t$\label{fig:MSDAlla}}%
\end{subfigure}%
\begin{subfigure}[b]{0.48\textwidth}%
	\centering%
	\includegraphics[width=\textwidth]{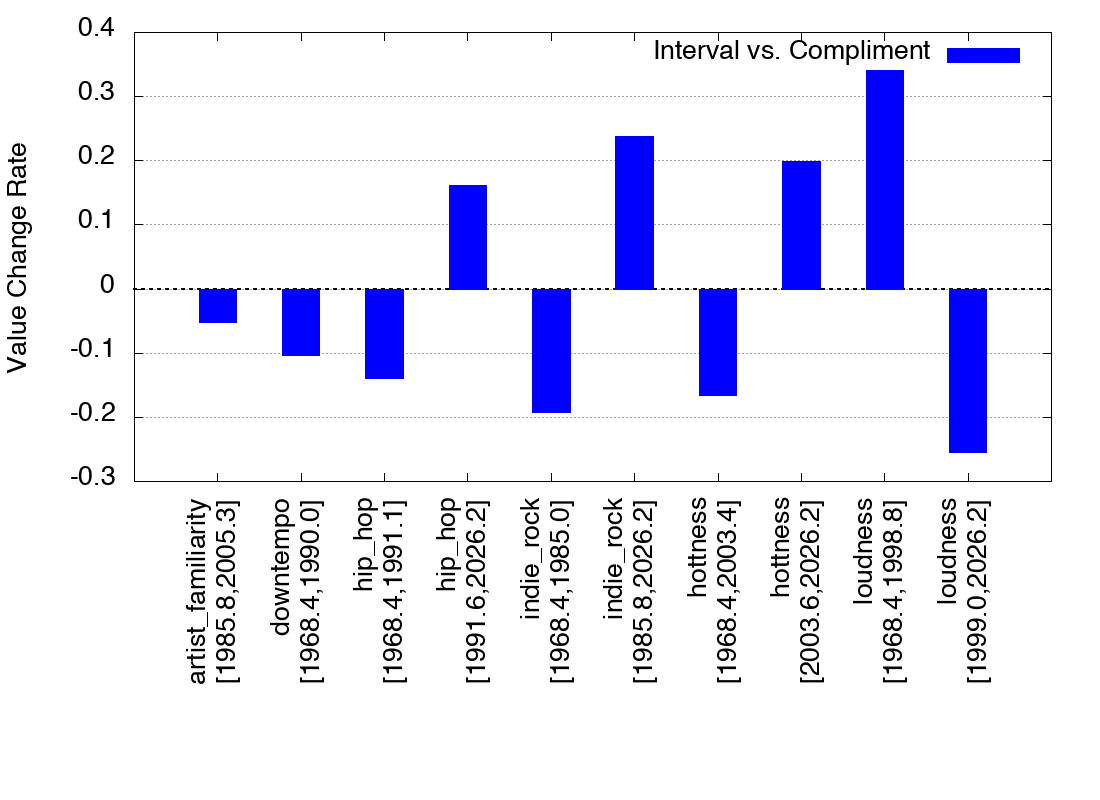}%
     \caption{Segments values vs. complement population values\label{fig:MSDAllb}}%
\end{subfigure}%
\caption{Top segments over all features.\label{fig:MSDall}}
\end{figure*}

\begin{figure*}[h!] 
\centering
 \captionsetup[subfigure]{aboveskip=-5pt,belowskip=-1pt}
\begin{subfigure}[b]{0.48\textwidth}
   \centering
	\includegraphics[width=\textwidth]{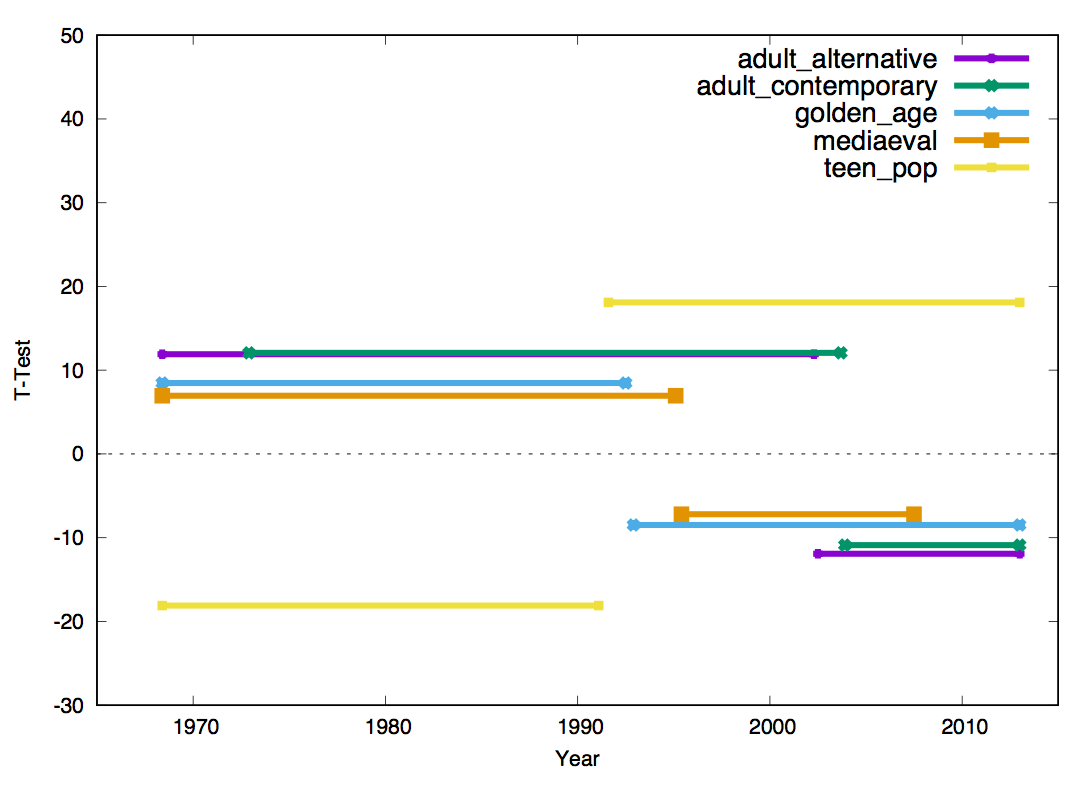}
  \vspace{0.005in}
        \caption{Target audience age\label{fig:MSDgroupa}}
\end{subfigure}
\begin{subfigure}[b]{0.48\textwidth}
	\centering
	\includegraphics[width=\textwidth]{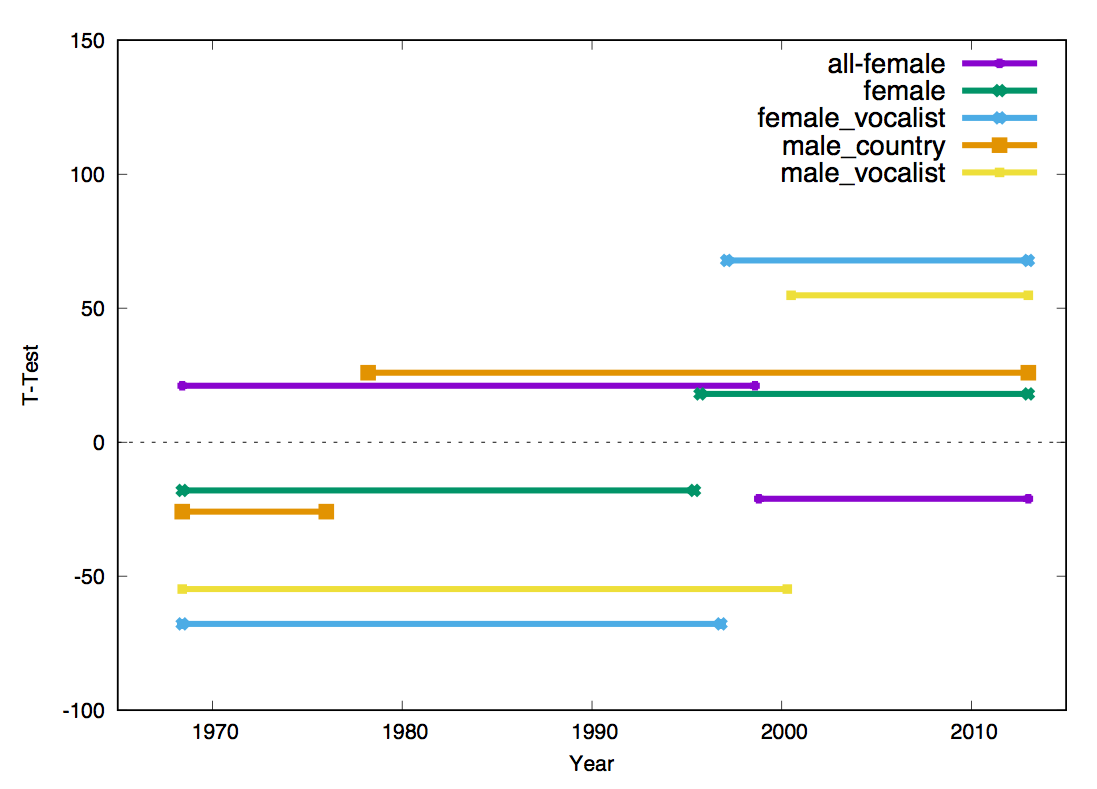}
   \vspace{0.005in}
	    \caption{Gender\label{fig:MSDgroupb}}
\end{subfigure}

\begin{subfigure}[b]{0.48\textwidth}
	\centering
	\includegraphics[width=\textwidth]{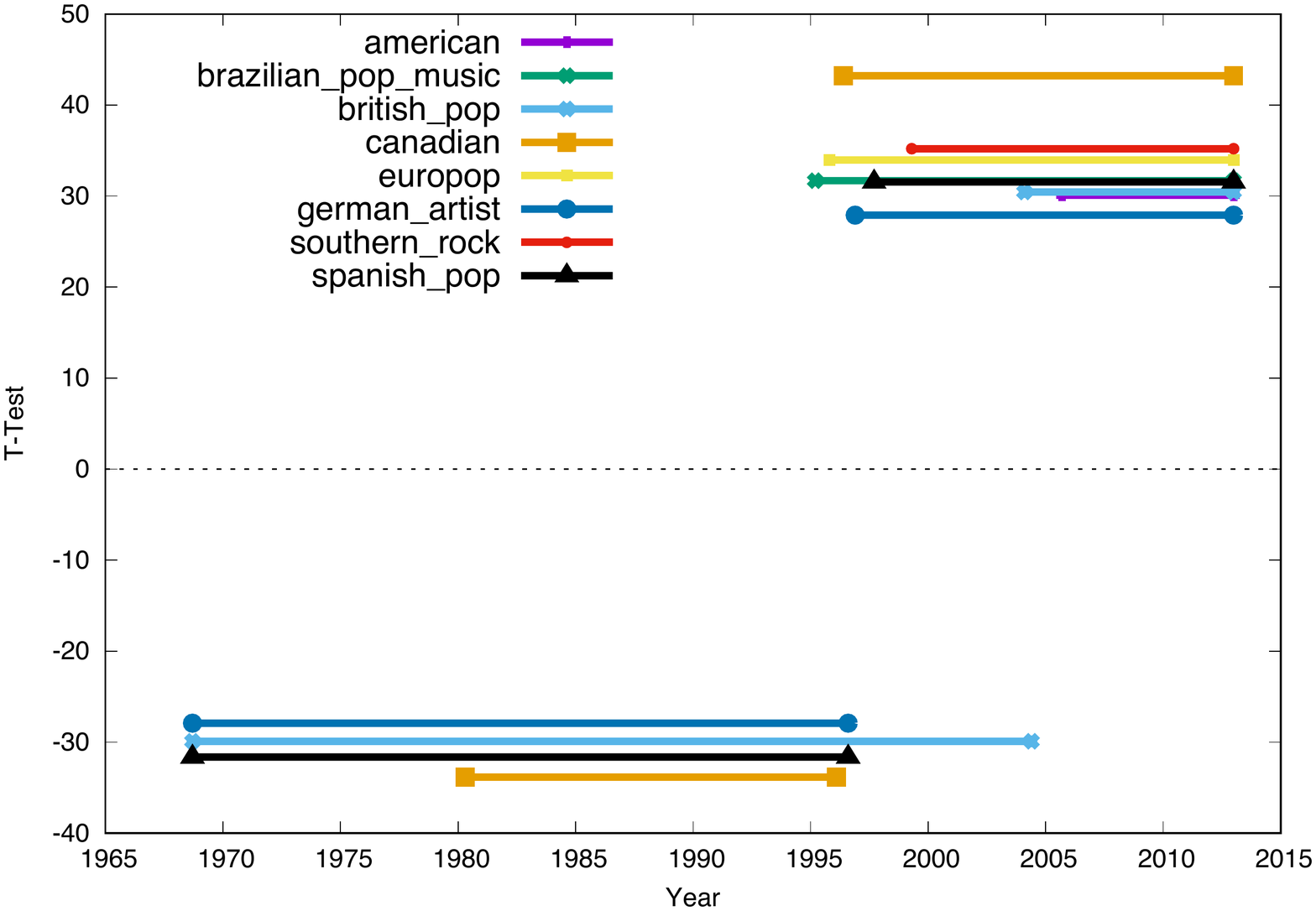}
        \caption{Location\label{fig:MSDgroupc}}
\end{subfigure}
 \begin{subfigure}[b]{0.48\textwidth}
 	\centering
 	\includegraphics[width=\textwidth]{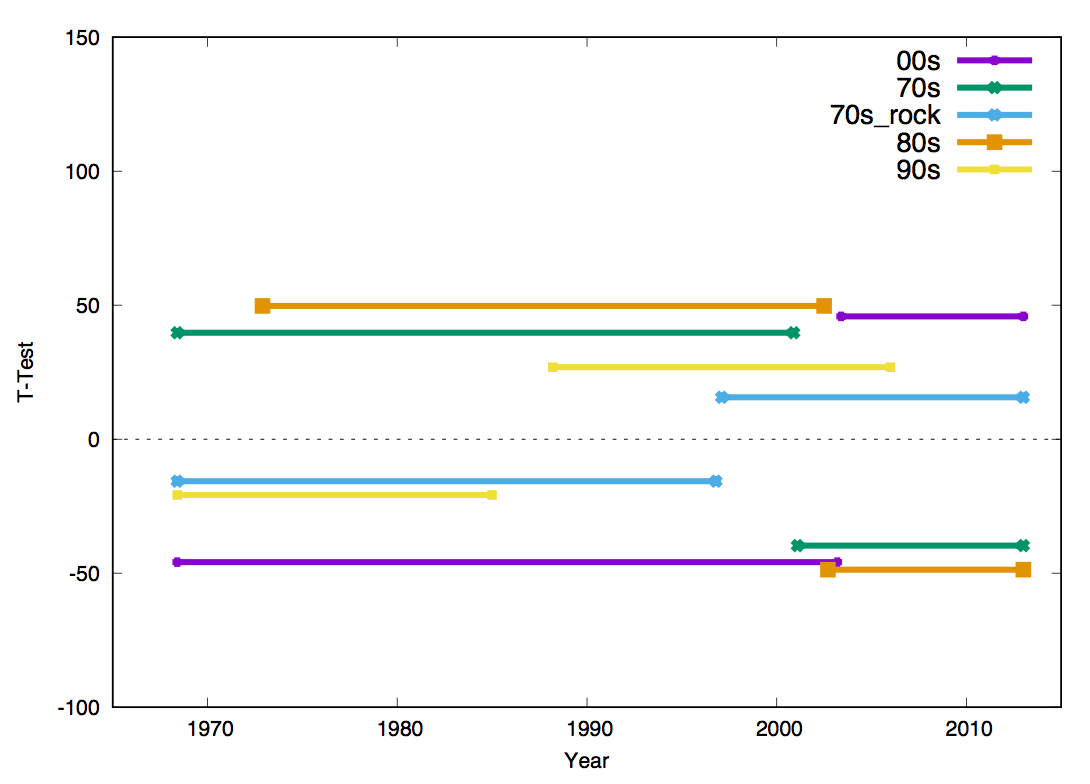}
     \vspace{0.005in}
         \caption{Musical decade\label{fig:MSDgroupd}}
  \end{subfigure}
 \caption{Most descriptive segments of different features groups.}
\label{fig:MSDgroup}
 \vspace{-0.1in}
\end{figure*}



\begin{figure*}[h!] 
\centering
\captionsetup{justification=centering}
\begin{subfigure}[b]{0.48\textwidth}%
	\centering
	\includegraphics[width=\textwidth]{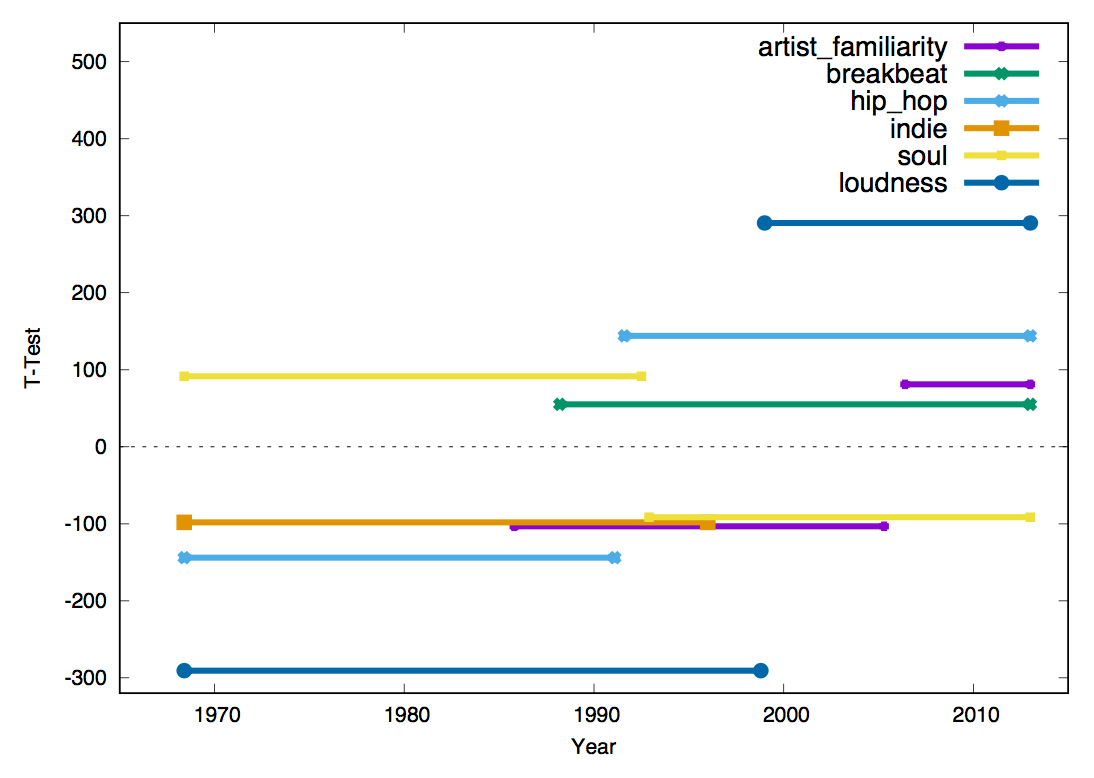}%
    \caption{One representative from each cluster\label{fig:MSDAlla}}%
\end{subfigure}%
\begin{subfigure}[b]{0.48\textwidth}%
	\centering%
	\includegraphics[width=\textwidth]{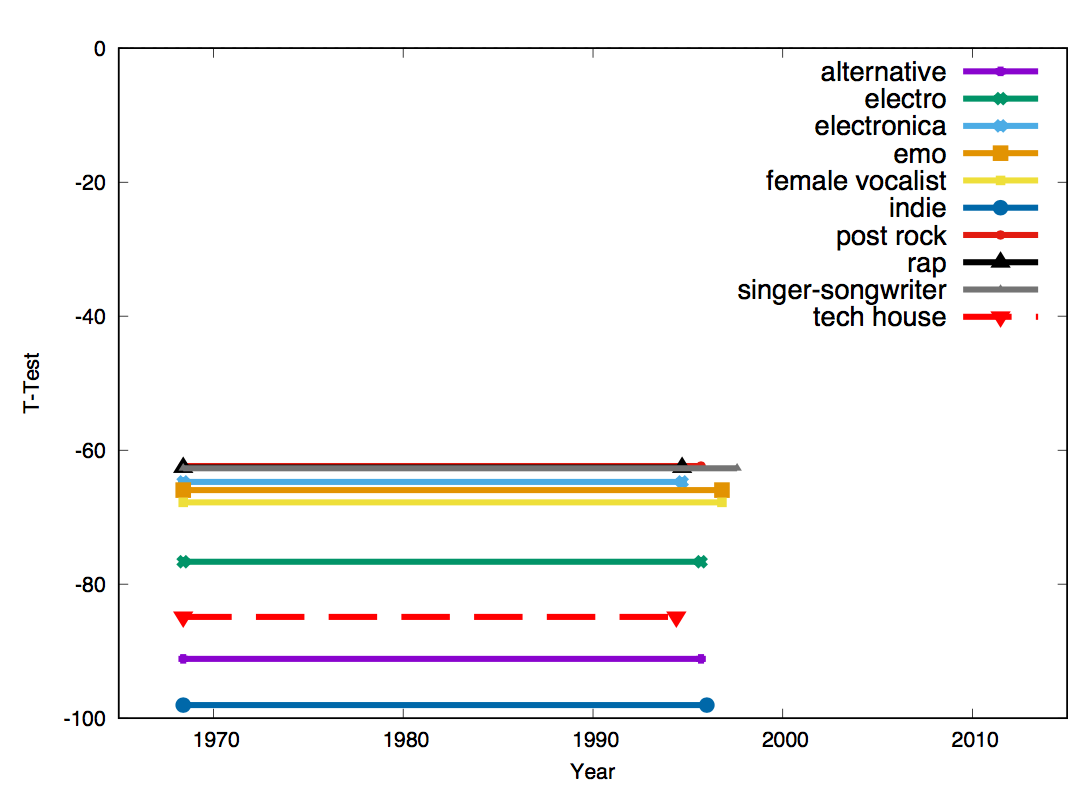}%
     \caption{'indie' cluster\label{fig:MSDAllb}}%
\end{subfigure}%

\begin{subfigure}[b]{0.48\textwidth}%
	\centering%
	\includegraphics[width=\textwidth]{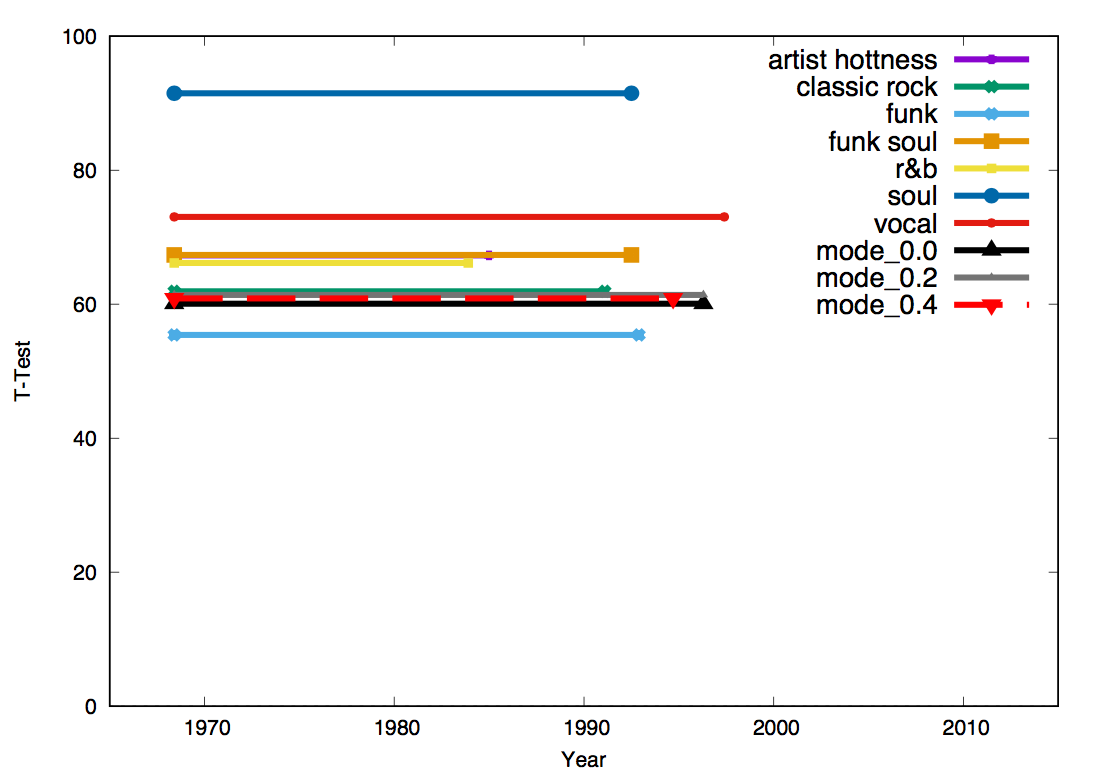}%
     \caption{'soul - early years' cluster\label{fig:MSDAllb}}%
\end{subfigure}%
\begin{subfigure}[b]{0.48\textwidth}%
	\centering%
	\includegraphics[width=\textwidth]{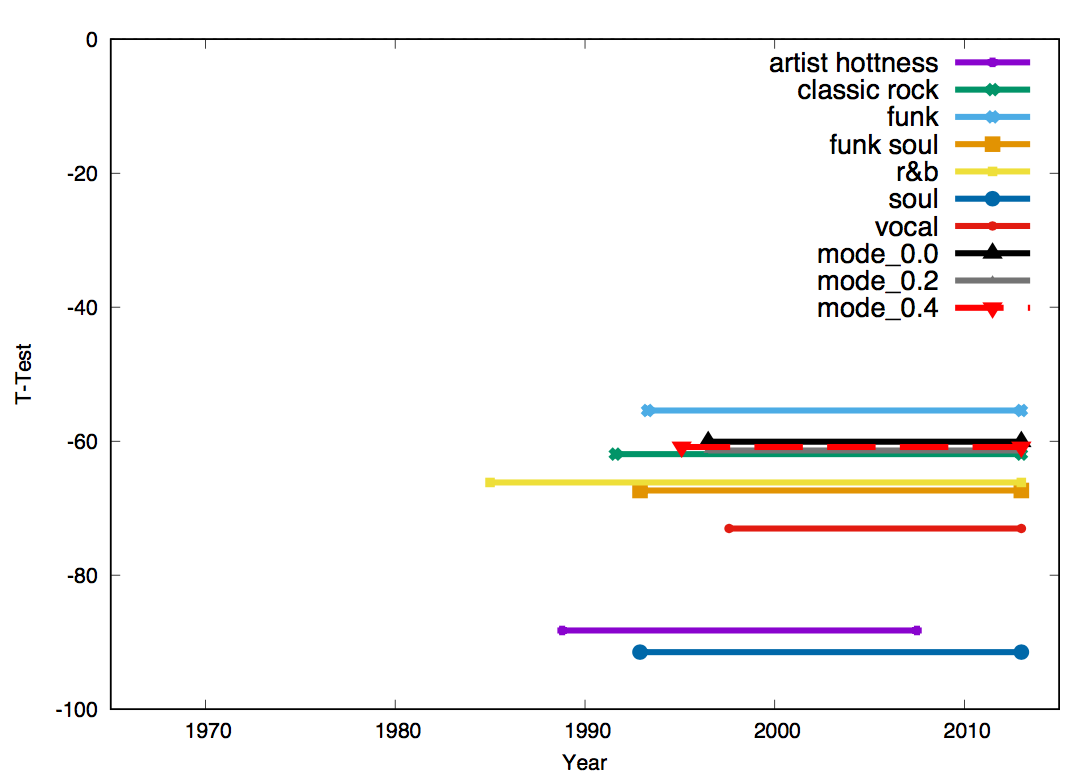}%
     \caption{'soul - late years' cluster\label{fig:MSDAllb}}%
\end{subfigure}%

\caption{Clustering of Segments\label{fig:MSD-Clusterall}}
\end{figure*}

Having trained and selected the model, we used the full set of one million songs to interpret it. The MSD dataset has 7,400 features which relate to different musical, vocal and geographic features of the song as well as genre and textual features. 

An example for one such feature is the "pop\_punk" genre indicator. Figure \ref{fig:MSD_pop_punk}(a) exemplifies the propensity of that feature for songs which were released in different years. As can be seen, the genre gained popularity starting in early 90's and became prominent after the turn of the millennium.  Figure \ref{fig:MSD_pop_punk}(b) depicts the dissimilarity of each of one hundred bins to the rest of the period. Figure \ref{fig:MSD_pop_punk}(c) shows that the maximal dissimilarity is found if years are segmented to those before and those after roughly 1996. 

Figure~\ref{fig:MSDall} depicts the most dissimilar feature segments in the release year prediction model. Some of those features are to be expected: more recent genre such as "hip\_hop" and "indie\_rock" describe songs which are predicted to be more recent. It is also not surprising that lower familiarity of the artist is indicative of an older release date. Still, it is educational to learn from our interpretation that loudness is the hallmark of songs which were released at and after the turn of the millennium. 

Finally, the ability to uncover hidden relations is further stressed when the user requests to interpret a model in terms of a specific set of features. 
As example, Figure~\ref{fig:MSDgroup} depicts the most significant features while concentrating on age, gender, location related features and on musical decades features.

As can be seen in Figure~\ref{fig:MSD-Clusterall}, the interpretation algorithm provides a diverse description of the release date of a song. If the artist is familiar, then the year is recent and if the song is load then it was probably released after the 90's. Soul music is a representative of pre-90's music and if we look at the cluster of likewise features (Figure~\ref{fig:MSD-Clusterall}(c)) we find other early genre such as classic rock, funk, and r\&b.

%% file: conclusions.tex
\section{Conclusions}
Over the past decade there is a clear trend withing the machine learning community towards ever more complex models. This trend is driven by breakthroughs in machine learning such as the advent of deep learning and by the exponential growth of both data and processing power. It seems that the age of simple, intelligible, models is over. However, that progress does not invalidate the need to understand the resulting model; a need which is rooted in the persisting functions of humans as designers and auditors of the learning process. The interpretation of complex machine learning models can be seen as the human-computer interaction (HCI) aspect of machine learning.

This work presents a concept and an algorithm for interpreting complex models. We expand the language of explanation to relations between populations of example and families of label distributions. We show that such explanations can be efficiently found from large data. We exemplify the meaningfulness of such explanations with two test cases: The generally available and well understood million-song database and an industrial problem from the mail serving domain.

In future work, we plan to develop explanations that are multivariate rather than univariate. We also plan to explore the value of different partitioning of the label space. For example, we can partition the label space by looking for features which have dissimilar distributions in a segment, below the segment, and above the segment. Alternatively, the label space  can be partitioned using a distributional distance metric - e.g., the greater the proportion of males in the population the less likely the distribution of labels to be like a specific distribution.

%% file: main.bbl

\begin{thebibliography}{00}


\ifx \showCODEN    \undefined \def \showCODEN     #1{\unskip}     \fi
\ifx \showDOI      \undefined \def \showDOI       #1{#1}\fi
\ifx \showISBNx    \undefined \def \showISBNx     #1{\unskip}     \fi
\ifx \showISBNxiii \undefined \def \showISBNxiii  #1{\unskip}     \fi
\ifx \showISSN     \undefined \def \showISSN      #1{\unskip}     \fi
\ifx \showLCCN     \undefined \def \showLCCN      #1{\unskip}     \fi
\ifx \shownote     \undefined \def \shownote      #1{#1}          \fi
\ifx \showarticletitle \undefined \def \showarticletitle #1{#1}   \fi
\ifx \showURL      \undefined \def \showURL       {\relax}        \fi
\providecommand\bibfield[2]{#2}
\providecommand\bibinfo[2]{#2}
\providecommand\natexlab[1]{#1}
\providecommand\showeprint[2][]{arXiv:#2}

\bibitem[\protect\citeauthoryear{Anderson, Antenucci, Bittorf, Burgess,
  Cafarella, Kumar, Niu, Park, R{\'e}, and Zhang}{Anderson
  et~al\mbox{.}}{2013}]%
        {anderson2013brainwash}
\bibfield{author}{\bibinfo{person}{Michael~R Anderson}, \bibinfo{person}{Dolan
  Antenucci}, \bibinfo{person}{Victor Bittorf}, \bibinfo{person}{Matthew
  Burgess}, \bibinfo{person}{Michael~J Cafarella}, \bibinfo{person}{Arun
  Kumar}, \bibinfo{person}{Feng Niu}, \bibinfo{person}{Yongjoo Park},
  \bibinfo{person}{Christopher R{\'e}}, {and} \bibinfo{person}{Ce Zhang}.}
  \bibinfo{year}{2013}\natexlab{}.
\newblock \showarticletitle{Brainwash: A Data System for Feature Engineering.}.
  In \bibinfo{booktitle}{{\em CIDR}}.
\newblock


\bibitem[\protect\citeauthoryear{Angelino, Larus-Stone, Alabi, Seltzer, and
  Rudin}{Angelino et~al\mbox{.}}{2017}]%
        {angelino2017learning}
\bibfield{author}{\bibinfo{person}{Elaine Angelino}, \bibinfo{person}{Nicholas
  Larus-Stone}, \bibinfo{person}{Daniel Alabi}, \bibinfo{person}{Margo
  Seltzer}, {and} \bibinfo{person}{Cynthia Rudin}.}
  \bibinfo{year}{2017}\natexlab{}.
\newblock \showarticletitle{Learning Certifiably Optimal Rule Lists}. In
  \bibinfo{booktitle}{{\em Proceedings of the 23rd ACM SIGKDD International
  Conference on Knowledge Discovery and Data Mining}}. ACM,
  \bibinfo{pages}{35--44}.
\newblock


\bibitem[\protect\citeauthoryear{Arthur and Vassilvitskii}{Arthur and
  Vassilvitskii}{2007}]%
        {kmeanplusplus2007}
\bibfield{author}{\bibinfo{person}{David Arthur} {and} \bibinfo{person}{Sergei
  Vassilvitskii}.} \bibinfo{year}{2007}\natexlab{}.
\newblock \showarticletitle{K-means++: The Advantages of Careful Seeding}. In
  \bibinfo{booktitle}{{\em Proceedings of the Eighteenth Annual ACM-SIAM
  Symposium on Discrete Algorithms}} {\em (\bibinfo{series}{SODA '07})}.
  \bibinfo{publisher}{Society for Industrial and Applied Mathematics},
  \bibinfo{address}{Philadelphia, PA, USA}, \bibinfo{pages}{1027--1035}.
\newblock
\showISBNx{978-0-898716-24-5}
\showURL{%
\url{http://dl.acm.org/citation.cfm?id=1283383.1283494}}


\bibitem[\protect\citeauthoryear{Bertin-Mahieux, Ellis, Whitman, and
  Lamere}{Bertin-Mahieux et~al\mbox{.}}{2011}]%
        {Bertin-Mahieux2011}
\bibfield{author}{\bibinfo{person}{Thierry Bertin-Mahieux},
  \bibinfo{person}{Daniel~P.W. Ellis}, \bibinfo{person}{Brian Whitman}, {and}
  \bibinfo{person}{Paul Lamere}.} \bibinfo{year}{2011}\natexlab{}.
\newblock \showarticletitle{The Million Song Dataset}. In
  \bibinfo{booktitle}{{\em {Proceedings of the 12th International Conference on
  Music Information Retrieval ({ISMIR} 2011)}}}.
\newblock


\bibitem[\protect\citeauthoryear{Biran and McKeown}{Biran and McKeown}{2017}]%
        {biran2017}
\bibfield{author}{\bibinfo{person}{Or Biran} {and} \bibinfo{person}{Katheleen
  McKeown}.} \bibinfo{year}{2017}\natexlab{}.
\newblock \showarticletitle{Human-Centric Justification of Machine Learning
  Predictions}. In \bibinfo{booktitle}{{\em Proceedings of the 26th
  International Joint Conference on Artificial Intelligence}} {\em
  (\bibinfo{series}{IJCAI '17})}.
\newblock


\bibitem[\protect\citeauthoryear{Bischof, Leonardis, and Selb}{Bischof
  et~al\mbox{.}}{1999}]%
        {BischofLS99}
\bibfield{author}{\bibinfo{person}{Horst Bischof}, \bibinfo{person}{Ales
  Leonardis}, {and} \bibinfo{person}{Alexander Selb}.}
  \bibinfo{year}{1999}\natexlab{}.
\newblock \showarticletitle{{MDL} Principle for Robust Vector Quantisation}.
\newblock \bibinfo{journal}{{\em Pattern Anal. Appl.\/}} \bibinfo{volume}{2},
  \bibinfo{number}{1} (\bibinfo{year}{1999}), \bibinfo{pages}{59--72}.
\newblock
\showDOI{%
\url{https://doi.org/10.1007/s100440050015}}


\bibitem[\protect\citeauthoryear{Chen and Guestrin}{Chen and Guestrin}{2016}]%
        {XGBoost}
\bibfield{author}{\bibinfo{person}{Tianqi Chen} {and} \bibinfo{person}{Carlos
  Guestrin}.} \bibinfo{year}{2016}\natexlab{}.
\newblock \showarticletitle{XGBoost: A Scalable Tree Boosting System}. In
  \bibinfo{booktitle}{{\em Proceedings of the 22Nd ACM SIGKDD International
  Conference on Knowledge Discovery and Data Mining}} {\em
  (\bibinfo{series}{KDD '16})}. \bibinfo{publisher}{ACM}, \bibinfo{address}{New
  York, NY, USA}, \bibinfo{pages}{785--794}.
\newblock
\showISBNx{978-1-4503-4232-2}
\showDOI{%
\url{https://doi.org/10.1145/2939672.2939785}}


\bibitem[\protect\citeauthoryear{Forgy}{Forgy}{1965}]%
        {forgy65}
\bibfield{author}{\bibinfo{person}{E. Forgy}.} \bibinfo{year}{1965}\natexlab{}.
\newblock \showarticletitle{Cluster Analysis of Multivariate Data: Efficiency
  versus Interpretability of Classification}.
\newblock \bibinfo{journal}{{\em Biometrics\/}} \bibinfo{volume}{21},
  \bibinfo{number}{3} (\bibinfo{year}{1965}), \bibinfo{pages}{768--769}.
\newblock


\bibitem[\protect\citeauthoryear{Hinkley}{Hinkley}{1971}]%
        {Hink71}
\bibfield{author}{\bibinfo{person}{D.V. Hinkley}.}
  \bibinfo{year}{1971}\natexlab{}.
\newblock \showarticletitle{Inference about the Change-Point from Cumulative
  Sum Tests}.
\newblock \bibinfo{journal}{{\em Biometrika\/}}  \bibinfo{volume}{58}
  (\bibinfo{year}{1971}), \bibinfo{pages}{509--523}.
\newblock
\showURL{%
\url{http://www.jstor.org/stable/2334386}}


\bibitem[\protect\citeauthoryear{Lipton}{Lipton}{2016}]%
        {ZCL2016Mythos}
\bibfield{author}{\bibinfo{person}{Zachary~C. Lipton}.}
  \bibinfo{year}{2016}\natexlab{}.
\newblock \showarticletitle{The Mythos of Model Interpretability}. In
  \bibinfo{booktitle}{{\em ICML Workshop on Human Interpretability of Machine
  Learning}}.
\newblock


\bibitem[\protect\citeauthoryear{Lloyd}{Lloyd}{1982}]%
        {lloyd82}
\bibfield{author}{\bibinfo{person}{Stuart~P. Lloyd}.}
  \bibinfo{year}{1982}\natexlab{}.
\newblock \showarticletitle{Least squares quantization in {PCM}}.
\newblock \bibinfo{journal}{{\em {IEEE} Trans. Information Theory\/}}
  \bibinfo{volume}{28}, \bibinfo{number}{2} (\bibinfo{year}{1982}),
  \bibinfo{pages}{129--136}.
\newblock
\showDOI{%
\url{https://doi.org/10.1109/TIT.1982.1056489}}


\bibitem[\protect\citeauthoryear{PAGE}{PAGE}{1954}]%
        {page54}
\bibfield{author}{\bibinfo{person}{E.~S. PAGE}.}
  \bibinfo{year}{1954}\natexlab{}.
\newblock \showarticletitle{Continuous inspection schemes}.
\newblock \bibinfo{journal}{{\em Biometrika\/}} \bibinfo{volume}{41},
  \bibinfo{number}{1--2} (\bibinfo{year}{1954}), \bibinfo{pages}{100--115}.
\newblock
\showDOI{%
\url{https://doi.org/10.1093/biomet/41.1-2.100}}


\bibitem[\protect\citeauthoryear{Ribeiro, Singh, and Guestrin}{Ribeiro
  et~al\mbox{.}}{2016}]%
        {RibeiroSG16}
\bibfield{author}{\bibinfo{person}{Marco~T{\'{u}}lio Ribeiro},
  \bibinfo{person}{Sameer Singh}, {and} \bibinfo{person}{Carlos Guestrin}.}
  \bibinfo{year}{2016}\natexlab{}.
\newblock \showarticletitle{"Why Should {I} Trust You?": Explaining the
  Predictions of Any Classifier}. In \bibinfo{booktitle}{{\em Proceedings of
  the 22nd {ACM} {SIGKDD} International Conference on Knowledge Discovery and
  Data Mining, San Francisco, CA, USA, August 13-17, 2016}}.
  \bibinfo{pages}{1135--1144}.
\newblock
\showDOI{%
\url{https://doi.org/10.1145/2939672.2939778}}


\bibitem[\protect\citeauthoryear{Robnik-\u{S}ikonja and
  Kononenko}{Robnik-\u{S}ikonja and Kononenko}{2008}]%
        {robnik2008}
\bibfield{author}{\bibinfo{person}{Marko Robnik-\u{S}ikonja} {and}
  \bibinfo{person}{Igor Kononenko}.} \bibinfo{year}{2008}\natexlab{}.
\newblock \showarticletitle{Explaining Classification for Individual
  Instances}.
\newblock \bibinfo{journal}{{\em IEEE Transactions on Knowledge and Data
  Engineering (TKDE)\/}}  \bibinfo{volume}{20} (\bibinfo{year}{2008}),
  \bibinfo{pages}{589--600}.
\newblock


\bibitem[\protect\citeauthoryear{Rudin}{Rudin}{2014}]%
        {Rudin:2014:AIM:2623330.2630823}
\bibfield{author}{\bibinfo{person}{Cynthia Rudin}.}
  \bibinfo{year}{2014}\natexlab{}.
\newblock \showarticletitle{Algorithms for Interpretable Machine Learning}. In
  \bibinfo{booktitle}{{\em Proceedings of the 20th ACM SIGKDD International
  Conference on Knowledge Discovery and Data Mining}} {\em
  (\bibinfo{series}{KDD '14})}. \bibinfo{publisher}{ACM}, \bibinfo{address}{New
  York, NY, USA}, \bibinfo{pages}{1519--1519}.
\newblock
\showISBNx{978-1-4503-2956-9}
\showDOI{%
\url{https://doi.org/10.1145/2623330.2630823}}


\bibitem[\protect\citeauthoryear{Saltelli, Chan, Scott, et~al\mbox{.}}{Saltelli
  et~al\mbox{.}}{2000}]%
        {saltelli2000sensitivity}
\bibfield{author}{\bibinfo{person}{Andrea Saltelli}, \bibinfo{person}{Karen
  Chan}, \bibinfo{person}{E~Marian Scott}, {et~al\mbox{.}}}
  \bibinfo{year}{2000}\natexlab{}.
\newblock \bibinfo{booktitle}{{\em Sensitivity analysis}}.
  Vol.~\bibinfo{volume}{1}.
\newblock \bibinfo{publisher}{Wiley New York}.
\newblock


\bibitem[\protect\citeauthoryear{Tolomei, Silvestri, Haines, and
  Lalmas}{Tolomei et~al\mbox{.}}{2017}]%
        {tolomei2017interpretable}
\bibfield{author}{\bibinfo{person}{Gabriele Tolomei}, \bibinfo{person}{Fabrizio
  Silvestri}, \bibinfo{person}{Andrew Haines}, {and} \bibinfo{person}{Mounia
  Lalmas}.} \bibinfo{year}{2017}\natexlab{}.
\newblock \showarticletitle{Interpretable Predictions of Tree-based Ensembles
  via Actionable Feature Tweaking}. In \bibinfo{booktitle}{{\em KDD}}.
\newblock


\bibitem[\protect\citeauthoryear{Wallace}{Wallace}{2017}]%
        {wallace17}
\bibfield{author}{\bibinfo{person}{Nick Wallace}.}
  \bibinfo{year}{2017}\natexlab{}.
\newblock \bibinfo{title}{EU's Right to Explanation: A Harmful Restriction on
  Artificial Intelligence}.
\newblock   (\bibinfo{year}{2017}).
\newblock


\bibitem[\protect\citeauthoryear{Ye, Chow, Chen, and Zheng}{Ye
  et~al\mbox{.}}{2009}]%
        {GBDT}
\bibfield{author}{\bibinfo{person}{Jerry Ye}, \bibinfo{person}{Jyh-Herng Chow},
  \bibinfo{person}{Jiang Chen}, {and} \bibinfo{person}{Zhaohui Zheng}.}
  \bibinfo{year}{2009}\natexlab{}.
\newblock \showarticletitle{Stochastic Gradient Boosted Distributed Decision
  Trees}. In \bibinfo{booktitle}{{\em Proceedings of the 18th ACM Conference on
  Information and Knowledge Management}} {\em (\bibinfo{series}{CIKM '09})}.
  \bibinfo{publisher}{ACM}, \bibinfo{address}{New York, NY, USA},
  \bibinfo{pages}{2061--2064}.
\newblock
\showISBNx{978-1-60558-512-3}
\showDOI{%
\url{https://doi.org/10.1145/1645953.1646301}}


\end{thebibliography}
